\documentclass{article} 
\usepackage{iclr2027_conference,times}

\usepackage{amsmath,amsfonts,bm}

\def\eqref#1{equation~\ref{#1}}

\def\1{\bm{1}}

\DeclareMathAlphabet{\mathsfit}{\encodingdefault}{\sfdefault}{m}{sl}
\SetMathAlphabet{\mathsfit}{bold}{\encodingdefault}{\sfdefault}{bx}{n}

\usepackage{hyperref}
\usepackage{url}
\usepackage{wrapfig}
\usepackage{caption}
\usepackage{booktabs}
\usepackage{graphicx}
\usepackage{amsmath}
\usepackage{graphicx}
\usepackage{subcaption}
\usepackage{algorithm}
\usepackage{algorithmic}
\usepackage{xspace}
\usepackage{booktabs}
\usepackage{amssymb,amsmath,amsthm,enumitem}
\usepackage{longtable}

\makeatletter
\newcommand\footnoteref[1]{\protected@xdef\@thefnmark{\ref{#1}}\@footnotemark}
\makeatother

\newcommand{\cbit}{\begin{compactitem}}
\newcommand{\ceit}{\end{compactitem}}
\newcommand{\cben}{\begin{compactenum}}
\newcommand{\ceen}{\end{compactenum}}

\newcommand{\beq}{\begin{equation}}
	\newcommand{\eeq}{\end{equation}}

\newcommand{\bit}{\begin{itemize}}
	\newcommand{\eit}{\end{itemize}}
\newcommand{\ben}{\begin{enumerate}}
	\newcommand{\een}{\end{enumerate}}

\newcounter{x}

  \newtheorem{definition}{Problem Statement}

\newcommand{\bx}{\mathbf{x}}

\newcommand{\bxaug}{\mathbf{x}_{\text{aug}}}
\newcommand{\yaug}{{y}_{\text{aug}}}

\definecolor{celadon}{rgb}{0.67, 0.88, 0.69}
\definecolor{carolinablue}{rgb}{0.6, 0.73, 0.89}

\newcommand{\Dtr}{\mathcal{D}_{\rm train}}
\newcommand{\Dts}{\mathcal{D}_{\rm test}}
\newcommand{\Dval}{\mathcal{D}_{\rm val}}

\definecolor{aliceblue}{rgb}{0.867, 0.917, 0.964}
\definecolor{aliceyellow}{rgb}{0.999, 0.945, 0.796}
\definecolor{alicegray}{rgb}{0.844, 0.867, 0.898}

\newcommand{\method}{{\sc VIP-COp}\xspace}
\iclrpreprinttrue

\title{\method: Black-box Context Optimization for Tabular Foundation Models}

\author{Yilong Chen \\
Carnegie Mellon University\\
\texttt{yilongch@andrew.cmu.edu} \\
\And
Xueying Ding \\
Carnegie Mellon University \\
\texttt{xding2@cs.cmu.edu} \\
\AND
Leman Akoglu \\
Carnegie Mellon University \\
\texttt{lakoglu@cs.cmu.edu}
}

\begin{document}

\maketitle

\begin{abstract}
Tabular foundation models (TFMs) have emerged as a powerful paradigm for in-context learning on structured data, enabling direct prediction on new tabular tasks without task-specific training. However, their effectiveness is constrained by context length limits, restricting application to medium-scale data and degrading performance when inference-time data exceed pretraining (size) distributions. Our work introduces VIP-COP, estimating the Value of Importance for Prediction of training examples and features for hard Context OPtimization for TFMs. Its explicit selection mechanism suppresses noise and isolates influential data, enabling the model to also benefit from data augmentation by prioritizing high-value augmented samples and features. VIP-COP is (i) fast, boosting performance often within minutes of optimization, based on an online KernelSHAP based regression with iterative refinement, value-guided context sampling, and multi-fidelity pruning; (ii) budget-aware and any-time, improving with additional test-time compute unlike heuristics that produce fixed context; and (iii) model-aware yet fully black-box, requiring no access to model internals, making it compatible with both proprietary and open-source TFMs; (iv) interpretable, identifying discrete “Very Important Predictors” (samples and features) that maximize signal-to-noise; which makes it (v) robust, isolating high-value data from noise. In contrast,soft-prompt optimization requires model gradients, produce abstract latent tokens,and lack explicit signal discrimination. Extensive experiments show that VIP-COP consistently outperforms heuristic and optimized baselines across large-scale high-dimensional testbeds, including data augmentation and data noise settings, establishing a new state of the art in test-time context refinement for TFMs.
\end{abstract}

\section{Introduction}

Machine learning (ML) has undergone a remarkable transformation with the advent of new model architectures and the rise of generative and foundation models that learn from massive datasets, exhibiting unprecedented generalization and in-context learning capabilities \citep{xie2021explanation,pmlr24icl}. This paradigm shift has permeated nearly all branches of AI, extending to tabular ML, where traditional methods are challenged by emerging Tabular Foundation Models (TFMs).

Pretrained TFMs aim to address classical problems \textit{in context} \citep{conf/icml/Nagler23}, including classification, regression, outlier detection, and density estimation \citep{hollmann2023tabpfn,hollmann2025accurate,qu2025tabicl,zhang2025mitra,shen2025fomod,pmlr26outformer}, where the training data is ingested as context samples and estimations on test data is output through a mere forward pass---without requiring any ML model training or model selection (i.e. hyperparameter tuning).

This shift creates an important question: \emph{which training examples and features should a TFM actually see?} This question arises naturally when the training data exceed the model's pretrained context capacity, but it remains important even when all available data fit within the context. Real-world datasets often contain irrelevant, noisy, corrupted, or redundant samples and features, making the full available context not necessarily optimal. Moreover, data augmentation, commonly used in practice to introduce additional variation, can inadvertently introduce further noise or uninformative examples \citep{journals/tmlr/YooZA23}. We refer to selecting a discrete subset of training examples and/or features for a frozen TFM as \emph{hard context optimization} (HardCOp).

More broadly, this challenge falls under prompt engineering or context optimization (COp) \citep{sahoo2024systematic,chen2025unleashing,yao2023visual,mei2025survey}. In NLP, COp typically focuses on prompts, retrieving relevant documents, or selecting few-shot examples to elicit desired behaviors. In the tabular domain, the ``instructions'' are implicit in the data.
Heuristic approaches to TFM context optimization aim to construct representative subsets or sketches of the data \citep{xu2025mixture,ye2025closer,feuer2023scaling}, which however, optimize \textit{model-agnostic} objectives.
In contrast, soft prompt optimization (SoftCOp) \citep{ma2024context,feuer2024tunetables} learn abstract context tokens that are optimized end-to-end to maximize performance over the full training set.

In this work, we introduce \method\footnote{\method is hosted at   {\footnotesize{\url{https://anonymous.4open.science/r/VIP-COp}}} for open-API TFMs.},  a fast, any-time algorithm that identifies Very Important Predictors; i.e., the most influential samples and features in the input. The key insight is that the value of a context item is context-dependent: an item may be useful in one context but redundant in another. We therefore estimate item contributions using an additive surrogate inspired by KernelSHAP \citep{shapley1953value,lundberg2017unified} and iteratively importance-sample candidate contexts to refine these estimates. Our main contributions are as follows:

\vspace{-0.05in}
\begin{itemize}[leftmargin=10pt, noitemsep, topsep=0pt]
\item \textbf{Black-box hard context optimization.}
We formulate TFM context selection as a discrete optimization problem and introduce \method, which optimizes explicit samples and features \emph{without modifying or accessing the internals of the pretrained backbone TFM}.

\item \textbf{Value-guided, anytime and any-size selection.}
\method estimates context-dependent item importance through an additive surrogate and importance sampling, producing a global ranking that supports flexible optimization budgets and context sizes.

\item \textbf{Robust and efficient context refinement.}
Extensive experiments show consistent gains under original, augmented, and noisy contexts, on classification and outlier detection tasks with a number of SOTA foundation models as backbone.

\item \textbf{Generalization and interpretability.}
We evaluate \method across TFM backbones, including TabPFN \citep{hollmann2023tabpfn}, TabICLv2 \citep{qu2026tabiclv2betterfasterscalable}, and Outformer \citep{pmlr26outformer}, and show that its sample/feature rankings provide a practical mechanism for context interpretation and data selection.

\end{itemize}

\vspace{-0.1in}
\section{Preliminaries}
\vspace{-0.05in}

\subsection{Tabular Foundation Models (TFMs)}
\label{ssec:tfms}
\vspace{-0.05in}

TFMs extend the foundation model paradigm to structured, tabular data by pretraining on synthetic tasks drawn from prior distributions over data-generating mechanisms. At inference time, a frozen TFM $q_{\boldsymbol{\theta}}$ takes a labeled training set $\Dtr=\{(\mathbf{x}_i,y_i)\}_{i=1}^{n}$ as \emph{context} and predicts labels for unlabeled \emph{query} points in a single forward pass, e.g.,
$q_{\boldsymbol{\theta}}(y \mid \mathbf{x}, \Dtr)$, without task-specific parameter optimization \citep{hollmann2023tabpfn}. We use $\Dval$ as a labeled query set for evaluating and selecting contexts, and $\Dts$ as the held-out test set for final evaluation. Thus, a candidate context $C \subseteq \Dtr$ contains a subset of context samples and features and is shared across query points. TabICL and TabICLv2 \citep{qu2025tabicl,qu2026tabiclv2betterfasterscalable} scale this mechanism to larger sample contexts via induced self-attention, while TabPFN-Wide \citep{kolberg2025tabpfn} scales to more features through feature partitioning and latent representations. OutFormer \citep{pmlr26outformer} extends the paradigm to outlier detection, where inlier examples form the context and query points are scored relative to the learned inlier structure. Despite these differences, TFM predictions depend directly on the samples and features included in the context, making context construction a natural target for optimization.

\subsection{Problem Statement and Settings}
\label{ssec}

\textbf{Context limits:}
Although Transformers can process arbitrary numbers of tokens architecturally, TFMs are pretrained on datasets within finite ranges of sample and feature sizes. Using substantially larger contexts at inference therefore constitutes a distribution shift and can degrade generalization. Context optimization thus serves two purposes: \textbf{(1) distillation:} reducing oversized datasets to pretrained limit, and \textbf{(2) denoising and efficiency:} removing irrelevant/misleading samples or features even when the data fit within the context. This not only improves the signal-to-noise ratio but also reduces the quadratic computational overhead of self-attention.

\begin{definition}[\textbf{Hard Context Optimization (HardCOp)}]
Given a frozen TFM $q_{\boldsymbol{\theta}}$ with context limits $(n_C,d_C)$ and $(\Dtr,\Dval,\Dts)$, select a global subset of $n' \leq n_C$ samples and $d' \leq d_C$ features from $\Dtr$ to maximize validation performance and generalize to $\Dts$.
\end{definition}
\vspace{-0.05in}

We optimize the context once and reuse it for all test queries, preserving efficient batched inference. We consider two settings, where same selection mechanism can both \textbf{exploit useful additional data} and \textbf{filter noisy or redundant information}.

\begin{definition}[\textbf{HardCOp under Data Augmentation (DA) \& Data Noise (DN)}] \textbf{HardCOp4DA:} $\Dtr$ is augmented with additional samples and/or features, potentially exceeding $(n_C,d_C)$, and HardCOp selects most informative subsets. \textbf{HardCOp4DN:} $\Dtr$ contains noisy samples and/or irrelevant features; HardCOp selects subsets that suppress harmful information.
\end{definition}

\subsection{Related Work}
\vspace{-0.025in}

\textbf{Context Optimization for Tabular Foundation Models (TFMs):}
Context optimization refines TFM inputs to improve predictions, analogous to prompt optimization for LLMs \citep{sahoo2024systematic}. 
Soft-prompt methods such as ICD \citep{ma2024context} and TuneTables \citep{feuer2024tunetables} compress training data into learned pseudo-points, but require white-box, gradient-based access; \method instead performs discrete, black-box selection over the original samples and features.
LoCalPFN \citep{thomas2024retrieval} and TabDPT \citep{ma2024tabdpt} retrieve nearest-neighbor contexts per query, while Xu \textit{et al.} \citep{xu2025mixture} route queries to learned local subsets. Unlike our global context, such query-specific adaptation inhibits efficient batching.
Other methods use model-agnostic sketching or dimensionality reduction \citep{feuer2023scaling}, or ensemble predictions over bootstrapped contexts \citep{liu2025tabpfn}, rather than explicitly optimizing a single context for the TFM.
Closest to ours, Rundel \textit{et al.} \citep{rundel2024interpretable} estimate sample and feature contributions using cached attention keys and values and prune low-value elements; unlike \method, this requires access to model internals.

\vspace{-0.025in}
\textbf{Scaling TFMs:}
Complementary work enlarges TFM capacity through architectural changes.
TabICL \citep{qu2025tabicl} uses Induced Self-Attention \citep{lee2019set}, reducing sample-wise attention from $O(N^2)$ to $O(NM)$ using $M$ inducing/router points \citep{shen2025fomod}, while TabPFN-Wide \citep{kolberg2025tabpfn} partitions features into blocks, reducing feature-wise attention from $O(P^2)$ to $O(PB)$.
In contrast, \method improves the context \emph{within} a fixed pretrained TFM rather than modifying its architecture.

\vspace{-0.025in}
\textbf{Long-context Adaptation:}
Long-context LLM methods similarly expand context capacity through position interpolation \citep{chen2023extending} or additional adaptation, as in LongLoRA and YaRN \citep{chen2023longlora,peng2023yarn}.
Unlike these approaches, \method requires neither architectural modification nor model fine-tuning or internal access.

\section{\method:  \underline{\textsc{V}}ery \underline{\textsc{I}}mportant \underline{\textsc{P}}redictors for \underline{\textsc{C}}ontext \underline{\textsc{Op}}timization}
\vspace{-0.05in}

\begin{figure}[t]
    \centering
    \includegraphics[width=\linewidth]{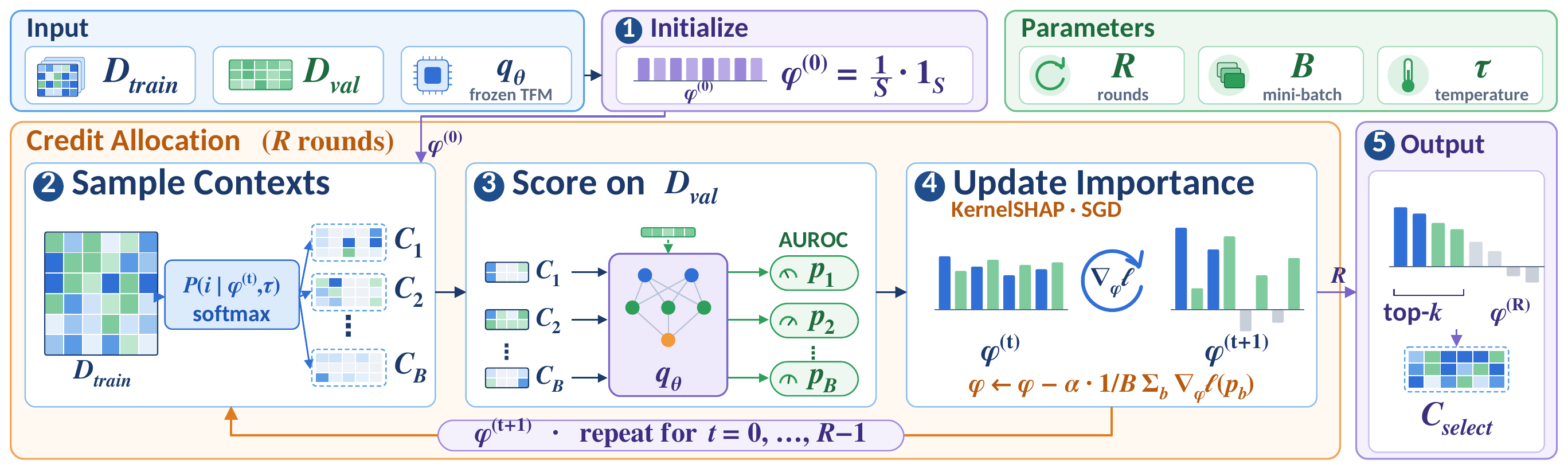}
    \caption{\method{} iteratively samples contexts from $\mathcal{D}_{\text{train}}$ via $\text{softmax}(\phi/\tau)$, scores them on $\mathcal{D}_{\text{val}}$ with the frozen TFM, and updates importance values $\phi$ by KernelSHAP-style SGD; after $R$ rounds, the top-$k$ items by $\phi$ form $C_{\text{select}}$.}
    \label{fig:algo}
    \vspace{-0.1in}
\end{figure}

Given as input ($i$) a labeled training set $\Dtr$ with $n$ examples and $d$ features, 
($ii$) a (small) labeled validation set $\mathcal{D}_{\mathrm{val}}$, and 
($iii$) a checkpoint $q_\theta$ of a TFM with a context capacity of $n_C$ examples and $d_C$ features, 
our proposed \method\ estimates the Values of Importance for Prediction (VIP) of the training examples as well as the feature dimensions whose cardinality exceeds the context limits.
Put differently, \method selects the examples and/or dimensions that are Very Important Predictors.

\vspace{-0.025in}
\textbf{Overview:~} 
TFMs act as ``algorithms-in-a-box,'' using in-context learning (ICL) to predict query examples from a training context in a single forward pass. Because the implicitly learned prediction rule changes with the context, each sample or feature has a context-dependent contribution. We therefore formulate hard context optimization as a \textbf{credit allocation problem} and estimate sample- and feature-level value using Shapley values \citep{shapley1953value}.

Specifically, the \textbf{collective value} of a context $\mathcal{C} \in \mathbb{R}^{n_c \times d_c}$, where $\mathcal{C} \subset \Dtr \in \mathbb{R}^{n \times d}$ with $n_c \le n$ and $d_c \le d$, is its validation performance when used to predict the query set $\mathcal{Q}:=\mathcal{D}_{\mathrm{val}}$. Shapley values aggregate an item's context-dependent marginal contributions across all possible subsets into a single value, which we call its VIP.

We approximate these values using Kernel SHAP \citep{lundberg2017unified}, formulating a linear regression where binary inputs $\mathbf{c}i=\mathbb{I}[i\in\mathcal{C}]\in{0,1};\forall i$ indicate item inclusion and validation performance $y{\mathrm{val}}$ is the response.\footnote{Original KernelSHAP performs a carefully weighted linear regression to accommodate varying size subsets. As we consider a \textit{fixed} size context (with $n_C$ rows and $d_C$ columns), we perform \textit{unweighted} least-squares.} The resulting coefficients approximate VIP and become exact when all possible subsets are enumerated.

Because exhaustive subset enumeration is infeasible, we iteratively refine VIP estimates by alternating between \textbf{(1)} updating the regression coefficients (VIP estimates) and \textbf{(2)} sampling new subsets according to the current VIP values. Each subset is evaluated on $\Dval$ with a single TFM forward pass, and its score is incorporated via mini-batch SGD. Temperature-guided importance sampling favors potentially high-VIP items while maintaining exploration, enabling efficient online refinement. Algorithm~\ref{alg:hardcop} summarizes \method (See visual overview in Fig. \ref{fig:algo}):

\renewcommand\algorithmicrequire{\textbf{Given:}}

\begin{algorithm}[t]
\scalebox{0.9}{
\begin{minipage}{\linewidth}
\caption{\method: \textbf{\textit{Hard Context Optimization for Black-Box Tabular Foundation Models}}}
\label{alg:hardcop}
\begin{algorithmic}[1]
\REQUIRE $\Dtr$, $(n,d)$$=$$\text{size}(\Dtr)$,  $\Dval$, TFM check-point $q_\theta$, $(n_C,d_C)$$=$$\text{Context\_size}(q_\theta)$; 
 \STATE \textbf{User-specified Input:} budget rounds $R$, $\eta$ (default $\eta=2$), mini-batch size $B$
\STATE \textbf{Initialization:} $r_{\max} = \lfloor \log_{\eta}(R)\rfloor$ \COMMENT{\#\textit{parallel runs}}, 
$S = n \times \mathbb{I}[n > n_C] + d \times \mathbb{I}[d > d_C]$,
$\boldsymbol{\phi}^{(0)} = \frac{1}{S}\,\mathbf{1}_S$ \COMMENT{\textit{init. all items' (examples and/or features) values}}

\FOR{$r \in 2\times \{r_{\max}, r_{\max}$$-$$1, \ldots, 0\}$ \COMMENT{\textit{parallel runs w/ varying temperature}}}
    \STATE temperature $\tau = \eta^{r-r_{\max}}$
    \FOR{$t \in \{0,1,\ldots, R$$-$$1\}$} 
    \STATE $P(i \mid \boldsymbol{\phi}^{(t)}, \tau)
= \frac{\exp\!\left(\phi_i^{(t)} / \tau\right)}
       {\sum_{j=1}^{S} \exp\!\left(\phi_j^{(t)} / \tau\right)} \;\; \forall i=1\ldots S$
       \FOR{$b=1\ldots B$ \COMMENT{\textit{parallel subset sampling and scoring}}} 
    \STATE $\mathcal{C}_b :=$
    {\texttt{sample} $(n_C, d_C)$ (examples, features) from $\Dtr$ w/ prob. $P(i \mid \boldsymbol{\phi}, \tau)$}
    \STATE $\widehat{\mathbf{y}}_{\text{val},b} :=$ \texttt{feed\_forward} $q_\theta(\Dval \;|\; \mathcal{C})$
    \STATE $p_b = $ \texttt{auroc}$(\widehat{\mathbf{y}}_{\text{val},b}, {\mathbf{y}}_{\text{val}})$ 
    \STATE $\mathbf{c}_b = \mathbf{1}_{[i \in \mathcal{C}_b]} \in \{0,1\}^S$
    \ENDFOR
    \STATE 
$\boldsymbol{\phi}^{(t+1)} \gets 
\boldsymbol{\phi}^{(t)} - \alpha \frac{1}{B} \sum_{b=1}^B 
\nabla_{\boldsymbol{\phi}} \ell\big(\boldsymbol{\phi}^{(t)}; \mathbf{c}_b, p_b\big)$ \COMMENT{\textit{mini-batch SGD update}}
\ENDFOR
\STATE $\mathcal{C}^{\text{select}}_r$$:=$top examples/features with the largest positive values $[\boldsymbol{\phi}^{(R)}]_+$
\STATE $\widehat{p}^{\text{select}}_{\text{val},r}  = \sum_{i \in \mathcal{C}^{\text{select}}_r} \boldsymbol{\phi}^{(R)}_i$
\ENDFOR
\RETURN  context $\mathcal{C}^{\text{select}}_r$ with the largest $\widehat{p}^{\text{select}}_{\text{val},r}$ 
\end{algorithmic}
\end{minipage}
}
\end{algorithm}\setlength{\textfloatsep}{0.2in}

\vspace{-0.025in}
\textbf{User-specified Input (Line 1):~} \method proceeds for a user-specified budget of $R$ rounds of mini-batch SGD updates with $B$ new importance-sampled subsets each. Our open source implementation takes around one second per round, so $R$ can be thought of as the time limit in seconds. $B$ is chosen based on available memory, where each subset (i.e., context) is drawn and fed-forward through the TFM to acquire validation performance in parallel.

\vspace{-0.015in}
\textbf{Initialization (Line 2):~}  
We optimize the context jointly over training examples and feature dimensions, yielding $S = n+d$ regression coefficients when both exceed context limits, and $n$ or $d$ coefficients otherwise. We initialize the coefficients uniformly by default;  more informative initializations include  proportional to attention-based scores \citep{hao2021self}, representative examples forming a coreset \citep{mirzasoleiman2020coresets} or features highly correlated with the target variable.

\vspace{-0.015in}
\textbf{Temperature-guided Importance Sampling (Lines 3--4)} plays a key role in our framework, as subsets are populated by selecting items according to their current VIP estimates, which may be unreliable in early rounds. If temperature $\tau$ is too low, sampling may prematurely concentrate on a few spuriously high-VIP items, limiting exploration and biasing the estimates. Conversely, if $\tau$ is too high, sampling explores too broadly, producing high-variance estimates under the finite budget of $R$$\times$$B$ subsets.

The appropriate temperature depends on the dataset, particularly the distribution of true VIP values and the quality of VIP initialization. Given a user-specified $\eta$ that controls available resources, we therefore run \method in parallel with $\lfloor \log_{\eta}(R) \rfloor$$+$$1$ different temperature schedules\footnote{This design is inspired by Hyperband \citep{li2018hyperband}, which terminates underperforming jobs at multiple training stages using multi-fidelity time-budget allocation to accommodate varying learning speeds.} (Lines 3--17), and return the best context across runs based on estimated validation performance (Line 18).

\vspace{-0.015in}
\textbf{Iterative VIP Refinement (Lines 5--14):~}
Over $R$ rounds (Line 5), we alternate between \textbf{(1)} enriching the regression with $B$ new subsets (Lines 7--12) and \textbf{(2)} updating the regression coefficients (VIP estimates) (Line 13). New subsets are constructed by importance-sampling items with probability proportional to their temperature-scaled VIP estimates (Line 6). The sampling temperature $\tau$, scheduled in Line 4, controls the trade-off between exploiting high-VIP items and exploring alternatives.

\vspace{-0.015in}
\textbf{Context Selection (Lines 15--18):~}
The linear regression yields an \textit{additive decomposition} of model performance, where the sum of VIP scores over a context equals its predicted validation performance. After $R$ rounds of SGD updates, we select the training examples and feature dimensions with the largest positive VIP estimates as the final context (Line 15). Their summed VIPs give the estimated validation performance $\widehat{p}{\text{val}}$ (Line 16). Finally, we return the context with the largest $\widehat{p}{\text{val}}$ across the independent temperature runs (Line 18).

\vspace{-0.05in}

\vspace{-0.05in}
\section{Evaluation}
\vspace{-0.1in}
\subsection{Experiment Setup}
\vspace{-0.015in}
\textbf{TFM Backbone:} We evaluate \method across a deliberately diverse set of state-of-the-art TFMs spanning different context scales and application settings. Specifically, we use \textbf{TabPFN-v1} \citep{hollmann2023tabpfn}, pretrained with contexts of up to $1{,}024$ samples and $100$ features, and \textbf{TabICLv2} \citep{qu2026tabiclv2betterfasterscalable}, which scales to substantially larger contexts of up to $48$K samples and $100$ features. We further evaluate \textbf{OutFormer} \citep{pmlr26outformer} for outlier detection, extending \method beyond standard classification tasks. Unlike classification, \textbf{OutFormer} uses clean inlier samples as context, providing a distinct evaluation setting with different context semantics.

\textbf{Datasets:~} For classification tasks, we source (38) multi-class classification datasets with $n$$>$$1,000$ from the TALENT\footnote{\url{https://github.com/LAMDA-Tabular/TALENT}} benchmark \citep{ye2025closer} and organize them into two: 
 ($i$) \textbf{small-d} includes 24 datasets with  $d $$<$$ 100$, of which 16 have $n $$<$$ 50K$ and 8  have $n $$>$$ 50K$; and 
($ii$) \textbf{Large-d} includes 14 datasets with $d $$>$$ 100$.   Appdx. \ref{app:datasets} and Table \ref{tab:datasets} provide further dataset details.
For outlier detection tasks, we source (240) outlier detection datasets with $n$$>$$6000$ from the MacrOData \footnote{\url{https://huggingface.co/spaces/MacrOData-CMU/MacrOData}} benchmark (specifically, OddBench) \citep{Ding_2026}. Appdx. \ref{app:datasets} and Table \ref{tab:odd_dataset_statistics} provide further dataset details. We use the standard train/val/test splits for all datasets.

\textbf{Baselines:~}
We compare \method with both heuristic \textbf{(H)} and optimized context-selection \textbf{(O)} strategies. All methods are evaluated in the same black-box TFM setting. We briefly describe baselines H1-H3 and O1-O2 as follows, with further details  in Appdx. \ref{app:baseconfigs}.

 \vspace{-0.015in}
\textit{Heuristic:}  While there are many possible model-unaware heuristics, we prioritize  common, easy to implement three: 
\textbf{(H1) Random} selects items (training examples and/or features, whichever exceeds context limits) uniformly at random. We report average test performance  across 15 random draws. 
\textbf{(H2) Ensemble} \citep{liu2025tabpfn} averages the predictions across 20 random contexts.
\textbf{(H3) XLContext} feeds $\Dtr$ fully as context without subsampling, potentially inducing out-of-(pretraining-)distribution context sizes.


\vspace{-0.015in}
\textit{Optimized:} 
\textbf{(O1) KMeansReps} fits a $n_C$-means clustering on the examples and selects the centroids as the context items for sample-level optimization.
For feature-level optimization, we instead perform $d_C$-means clustering on the features and select the centroids as the context features.
Validation performance is used to select the best clustering  among those trained
by 10 randomly initialized centroids.
\textbf{(O2) DT+TFM}  As baselined in \citep{ye2025closer}, a shallow decision tree with minimum leaf size 
$n_{\min} $$=$$ n_C$
 is trained to partition $\Dtr$ into subsets. Each validation example is routed to its corresponding leaf, where TFM uses the leaf's examples as context to produce the prediction.
We modify DT+TFM baseline for feature-level optimization by training a DT with maximum depth $d_{\max}$$=$$50$ and selecting up to 100 unique split features level-by-level starting from the root. 
Validation performance is used to select the best DT among those trained by choosing a random feature at each split from among the top-three as ranked by information gain in both scenarios. \emph{Note: this baseline is not viable under outlier detection settings since only inlier exists in the training data.}

\textbf{Metrics:~} 
We evaluate methods on tabular multi-class classification datasets w.r.t. \textit{balanced} accuracy and relative improvement (\%) over baselines; and methods on tabular outlier detection datasets w.r.t. {AUROC} and  improvement (\%) over baselines. Besides performance, we use paired permutation tests to assess significance of differences across datasets and report optimization time to quantify the test-time cost of context selection.

\vspace{-0.015in}
\subsection{Evaluation Settings}
\vspace{-0.015in}
\textbf{Data Augmentation (DA):~} We enrich each $\Dtr$ with augmented samples and/or features prior to context optimization \citep{bengio2017deep,Ericsson22Why}. Specifically, datasets with $n<50$K and $d<100$ are augmented to $n=50$K and $d=100$; those with $n>50$K and $d<100$ are feature-augmented only to $d=100$; and those with $n<50$K and $d>100$ are sample-augmented only to $n=50$K. For sample augmentation \textbf{($\text{S}_{\text{aug}}$)}, inspired by SMOTE \citep{Chawla_2002}, we generate $\bxaug=\alpha\bx_k+(1-\alpha)\bx_l$ from randomly sampled ${\bx_k,\bx_l}\in\Dtr$ with $\alpha\sim\mathcal{U}(0,1)$, assigning $\yaug=y_k$ if $\alpha\leq0.5$ and $\yaug=y_l$ otherwise. For feature augmentation \textbf{($\text{F}_{\text{aug}}$)}, we apply random projections of the form $\mathbf{X}_{\text{train}} \cdot \mathbf{R}$ to create $d_{\text{aug}}$ new features, where $\mathbf{X}_{\text{train}} \in \mathbb{R}^{n \times d}$  and  $\mathbf{R} \in \mathbb{R}^{d \times d_{\text{aug}}}$ with $\mathbf{R}_{ij} \sim \mathcal{N}(0,1/d_{\text{aug}})$. \textit{\textbf{Remark:}} augmentation choice is itself a model-selection problem, as different augmentations may be useful or irrelevant depending on the dataset \citep{Cubuk19Auto,zoph2020learning,ottoni2023tuning}. Rather than optimizing augmentation policies, which is beyond our scope, we use standard off-the-shelf augmentations as a proof of concept. Thus, our results provide a conservative estimate of the potential gains from better-designed augmentations, while showing that \method can mitigate harmful augmentations by filtering uninformative examples.

\textbf{Data Noise (DN):~} To stress-test robustness to irrelevant or corrupted context, we perturb the training data while preserving its original size. For small-$d$ datasets, we randomly drop half of the samples and replace them with noisy samples to recover the original $n$; for large-$d$ datasets, we randomly drop half of the features and replace them with noisy features to recover the original $d$. We consider two sample-noise strategies: \textbf{($\text{S1}_{\text{noi}}$)} samples each feature independently from its empirical marginal, and \textbf{($\text{S2}_{\text{noi}}$)} draws samples from a global Gaussian whose mean vector and covariance matrix are estimated from the original data. Both assign labels sampled from the empirical label distribution. For feature noise, \textbf{($\text{F1}_{\text{noi}}$)} copies a randomly chosen existing feature and adds Gaussian noise with three times its original variance, while \textbf{($\text{F2}_{\text{noi}}$)} randomly permutes the values of an existing feature across samples. \textit{\textbf{Remark:}} DN is designed as a controlled robustness stress test rather than a dataset-specific model of realistic corruption: constructing realistic noise would require knowledge of each dataset's underlying data-generating process, which is impractical at our study scale. Importantly, most baselines degrade substantially under DN, indicating that the setting is sufficiently challenging, whereas \method remains comparatively robust by filtering irrelevant or corrupted context.

\textbf{Outlier Detection (OD):~} To evaluate whether \method generalizes beyond standard supervised classification, we consider OutFormer \citep{pmlr26outformer}, a TFM designed for semi-supervised outlier detection with context size 5K including the context and the query/test sets in up to 100 features. It is pretrained with clean context consisting of inlier examples only, while test points are evaluated for anomalousness relative to the learned inlier structure. 

\textbf{Corruption Sensitivity (CS):~} We hypothesize that real inlier pools may contain outliers mislabeled as inliers. Since such label errors are unobservable, we mimic them by replacing randomly selected inliers with real, held-out outliers and treating them as inliers, creating data-derived contamination. We retain 83 OddBench datasets with at least $6$K inliers and $1{,}200$ labeled outliers, and construct disjoint $6$K-point context pools, $1$K query sets, and $5$K test sets. We replace $\varepsilon n_{\text{pool}}$  of randomly selected inlier points with outliers absent from both evaluation splits, using $\varepsilon\in\{0,0.05,0.1,0.2\}$. We then run \method{} with true query labels for $100$ rounds of $2{,}000$ coalitions of size $500$ and retain the top $k=3{,}000$ points by $\phi$. Keeping the pool and evaluation sets fixed across $\varepsilon$ isolates the effect of contamination. This setting creates more organic noise than the Data Noise (DN) setting.

\vspace{-0.05in}
\subsection{Main Results}

\begin{wrapfigure}{r}{0.425\textwidth}
    \centering
    \vspace{-0.15in}
    
    \includegraphics[
        width=0.97\linewidth
    ]{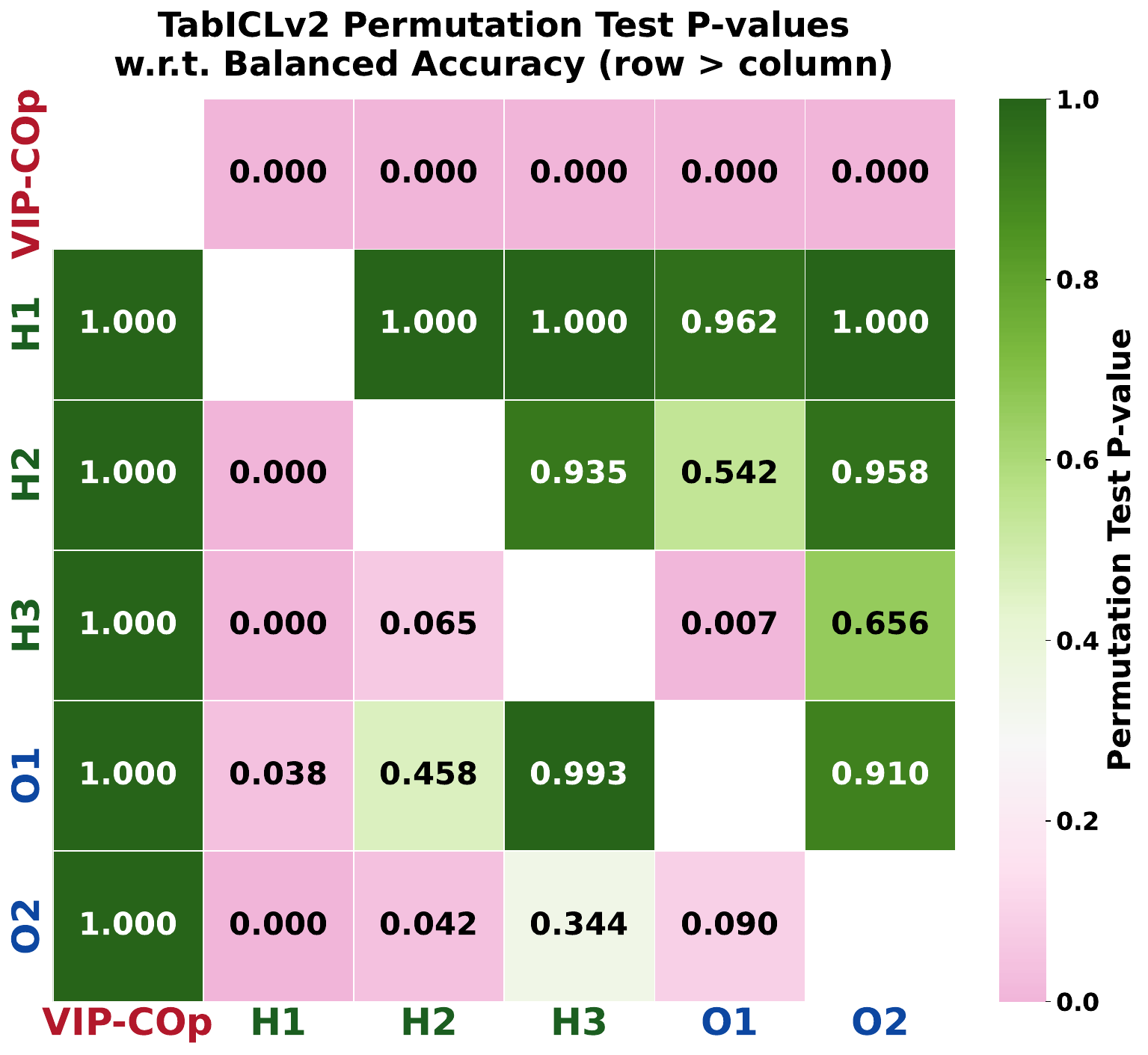}
    
    \vspace{-0.08in}
    \caption{
    Paired comparisons between all context optimization methods based on
    $p$-values of permutation tests (TabICLv2 as backbone).
    TabPFNv1 results in Fig.~\ref{fig:allpairs_v1}.
    }
    \label{fig:allpairs_v2}
    
    \vspace{-0.2in}
\end{wrapfigure}

\vspace{-0.015in}
\textbf{Originial Classification:~}
With the TabPFN-v1 as backbone and original classification tasks, \method consistently improves over the evaluated context-selection baselines across the 38 datasets providing 1.54\% to 14.68\% performance improvement; full per-dataset results available in Table \ref{tab:og_results_tab} of Appdx. \ref{app:results} .  
Further evaluation on the larger-context TabICLv2 backbone demonstrates that \method outperforms all five baselines (see Table \ref{tab:TabICL_OG_results_tab} in Appdx. \ref{app:results} for full results). Paired statistical tests in Fig. \ref{fig:allpairs_v2} and Table \ref{tab:pvals_v2_intext} confirm this finding, showing that \method achieves significant gains over every baseline ($p$$<$$0.01$). Although XLContext (H3) and DT+FM (O2) emerge as effective heuristics with similar performance to one another ($p$$=$$0.344$), both ultimately remain inferior to \method. Similar results on TabPFNv1 are in Appdx. Table~\ref{tab:pvals_v1} and Fig. \ref{fig:allpairs_v1}. 

\textbf{Data Augmentation (HardCOp4DA):~}
For TabPFNv1 (see details in  Appdx.~\ref{app:results}, Table~\ref{tab:aug_sample_results_tab} and Table~\ref{tab:aug_feature_results_tab}), we report individual performances under sample and feature augmentation, respectively. Across the HardCOp4DA experiments, \method improves over the baselines by 4.80\% to 15.05\%. Table~\ref{tab:pvals_v1} shows that \method outperforms all baselines across the data augmentation settings except one, where the difference between KMeansReps and \method is not statistically significant ($p=0.277$).

For TabICLv2 (see details in Appdx.~\ref{app:results}, Tables~\ref{tab:v2_aug_small_d}), \method improves over the baselines by approximately 1.9\%--5.8\%, with $p$$<$$0.01$ (see Table \ref{tab:pvals_v2_intext}). These results indicate that explicit context selection can exploit an expanded context without assuming that every augmented item is informative. Tables \ref{tab:aug_sample_results_tab}
and \ref{tab:aug_feature_results_tab} show that \method benefits particularly from sample augmentation as compared to feature augmentation, which motivate further research into data augmentation strategies for TFMs.

\vspace{0.1em}
\begin{wrapfigure}{r}{0.48\textwidth}
    \centering
    \vspace{-0.15in}
    
    \captionof{table}{
    Pairwise comparison between each baseline and \method{} 
    (TabICLv2 as backbone) based on $p$-values of permutation tests 
    over datasets across all HardCOp settings 
    (TabPFNv1 results in Appdx.~\ref{tab:pvals_v1}).
    }
    \label{tab:pvals_v2_intext}
    
    \vspace{-0.05in}
    \scalebox{0.73}{
    \begin{tabular}{lccc|cc}
        \toprule
         & Rand & Ens & XL & KRep & DT+FM \\
         & \textbf{(H1)} & \textbf{(H2)} & \textbf{(H3)}
         & \textbf{(O1)} & \textbf{(O2)} \\
        \midrule
        Original
        & $<$0.01 & $<$0.01 & $<$0.01 & $<$0.01 & $<$0.01 \\
        \midrule
        DA w/ $\text{S}_{\text{aug}}$
        & $<$1e-3 & $<$1e-3 & $<$1e-3 & $<$1e-3 & $<$1e-3 \\
        \midrule
        DN w/ $\text{S}_{\text{noi}}$
        & $<$1e-6 & $<$1e-6 & $<$1e-6 & $<$1e-6 & $<$1e-6 \\
        DN w/ $\text{F}_{\text{noi}}$
        & $<$0.01 & $<$0.01 & $<$0.01 & $<$0.01 & $<$0.01 \\
        \midrule
        \textbf{Overall}
        & \textbf{$<$1e-6}
        & \textbf{$<$1e-6}
        & \textbf{$<$1e-6}
        & \textbf{$<$1e-6}
        & \textbf{$<$1e-6} \\
        \bottomrule
    \end{tabular}
    }
    
    \vspace{-0.2in}
\end{wrapfigure}

\textbf{Data Noise (HardCOp4DN):~}
We next introduce controlled sample- and feature-level noise to reduce the signal-to-noise ratio of the context. For TabPFNv1,
(see details in Appdx.~\ref{app:results} and full results in Table~\ref{tab:noise_s1_results_tab}, Table \ref{tab:noise_s2_results_tab}, and Table \ref{tab:noise_f1f2_results_tab}) reports detailed results under the three data noising schemes. \method yields consistent improvements over the baselines, ranging from 1.92\% to 27.38\% under sample noise and from 154.11\% to 240.81\% under feature noise. 

Across the 24 TabICLv2 datasets, \method improves over the baselines by approximately 7.0--8.7\%, with all aggregate comparisons significant at $p<0.001$. This supports the use of explicit selection to suppress samples and features that are irrelevant or harmful to prediction. Table~\ref{tab:noise_s1_results_tab}, Table \ref{tab:noise_s2_results_tab}, and Table \ref{tab:noise_f1f2_results_tab} indicate that a lower signal-to-noise ratio degrades the performance of all methods, with the baselines experiencing more pronounced deterioration, particularly under feature noise.





\begin{figure*}[htbp]
\vspace{-0.1in}
    \centering
    \setlength{\tabcolsep}{2pt}
    \begin{tabular}{cc}
        \multicolumn{2}{c}{
            \includegraphics[width=0.77\linewidth]{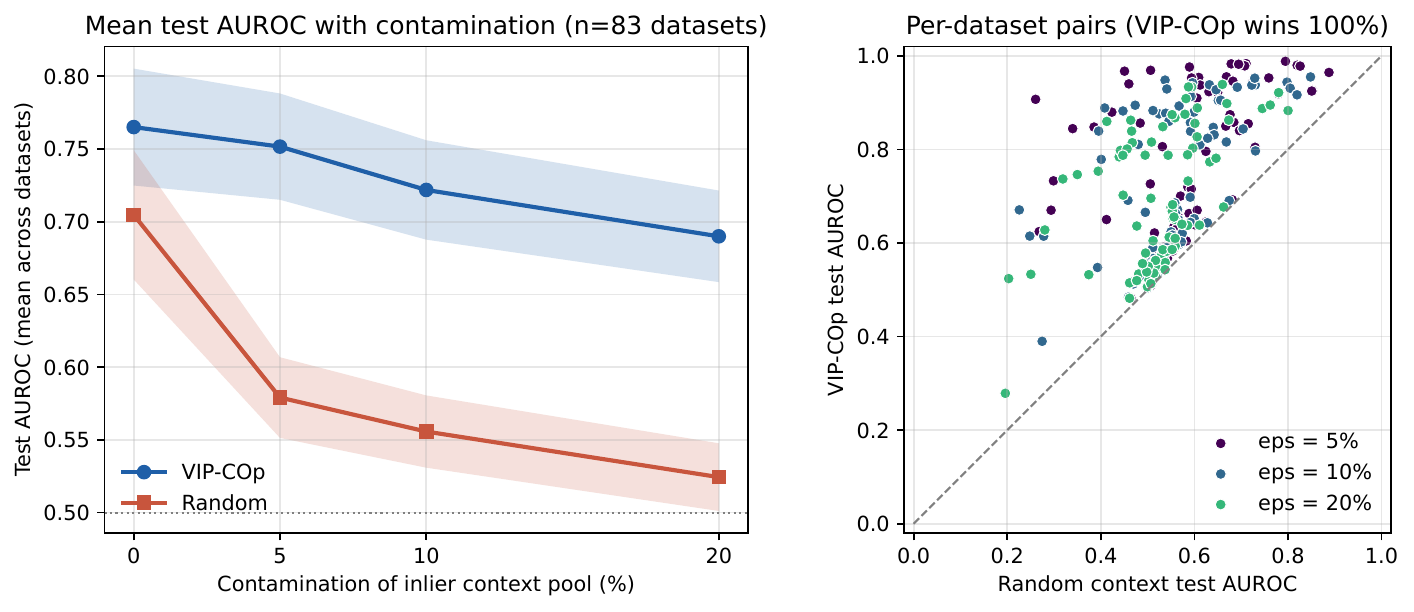}
        } \\[-0.2em]
        \includegraphics[width=0.4\linewidth]{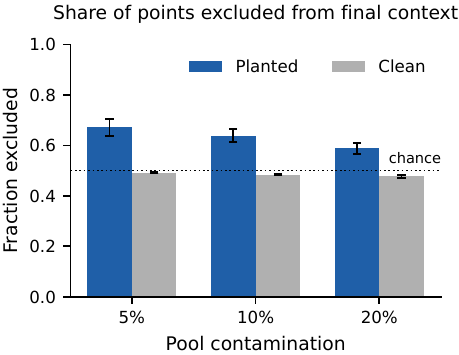} &
        \includegraphics[width=0.4\linewidth]{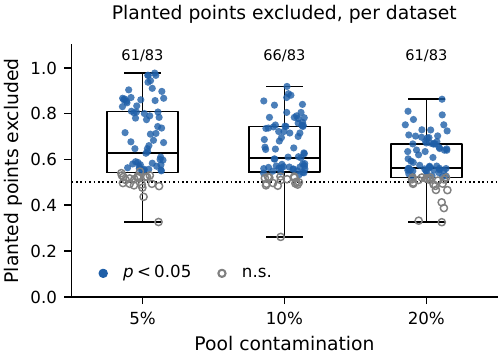}
    \end{tabular}
    \vspace{-0.1in}
    \caption{\textbf{\method{} degrades slower under context-pool corruption and preferentially excludes planted outliers.}
Held-out real outliers are planted into the inlier-only context pool of $83$ ODDBench datasets, disguised as inliers: \textbf{(a,b)} a random context collapses toward chance as contamination rises while \method{} largely holds, winning on every dataset at every rate; \textbf{(c,d)} clean points are excluded from the final context at the chance rate while planted points are excluded well above it, though the enrichment weakens as contamination grows.}
    \label{fig:contain_grid}
    \vspace{-0.1in}
\end{figure*}

\textbf{Outlier Detection:~} Beyond classification, on 240 outlier detection tasks from Oddbench using the Outformer \citep{pmlr26outformer} backbone,  \method also improves over all the five baselines by approximately 9.13\%-23.75\% (full per-dataset results available in Appdx. \ref{app:results} Table \ref{tab:odd_results_tab}). These results demonstrate that context optimization extends to TFMs with substantially different objectives and inference settings, including semi-supervised outlier detection.


\textbf{Robustness to Context Contamination:~}
We further test whether \method remains effective when the context pool is contaminated with outliers. Across 83 OddBench datasets, we replace $\epsilon \in {5\%,10\%,20\%}$ of the 6{,}000-point context pool with held-out outliers, from which \method selects 3{,}000 points. As shown in Fig.~\ref{fig:contain_grid}a, random-context AUROC drops from 0.705 on the clean pool to 0.579 and 0.525 at 5\% and 20\% contamination, whereas \method declines from 0.765 to 0.752, 0.722, and 0.690, widening its advantage from 0.06 to 0.17 AUROC (Fig.~\ref{fig:contain_grid}b). \method also excludes 67\%, 64\%, and 59\% of planted outliers, compared with $\sim$49\% of clean points (shown in Fig.~\ref{fig:contain_grid}c); this filtering is significant ($p<0.05$) on 61, 66, and 61 datasets, respectively (Fig.~\ref{fig:contain_grid}d).

\begin{figure}[htbp]
\label{CD_Diag}
\vspace{-0.1in}
\centering

\begin{subfigure}[t]{0.49\linewidth}
    \centering
    \includegraphics[width=\linewidth]{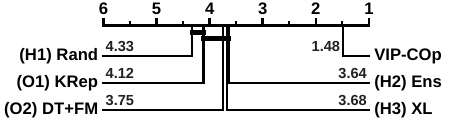}
    \label{fig:overalldemsar}
\end{subfigure}
\hfill
\begin{subfigure}[t]{0.49\linewidth}
    \centering
    \includegraphics[width=\linewidth]{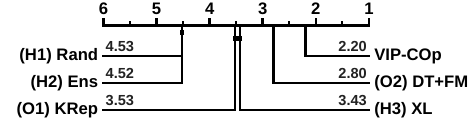}
    \label{fig:secondfigure}
\end{subfigure}
\vspace{-0.17in}
\caption{CD diagrams comparing methods across all settings with the TabICLv2 (left) and TabPFNv1 (right). With both backbones, \method{} achieves the best average rank and significantly outperforms all baselines. The baseline ordering depends on the backbone: the baselines are largely indistinguishable with TabICLv2. But \method{} outperforms with both backbones. }
\label{fig:overall_comparison}
\vspace{-0.1in}
\end{figure}
\textbf{Summary of Results:~} Fig.~\ref{fig:overall_comparison} reveals a consistent advantage for \method across diverse context-selection settings and model backbones. Despite substantial variation in the relative performance of the baselines across backbones, \textbf{\method consistently remains the top-ranked method, indicating that its gains are not tied to a particular TFM architecture.}
\vspace{-0.015in}
\subsection{Ablation Study and Efficiency Analysis}
\vspace{-0.015in}
\textbf{Compress Beyond the Context Limit:~} We examine the context size--performance trade-off by truncating \method's global data-point ranking after optimization. Using TabICLv2, we compare ranking prefixes with a context containing all samples with positive estimated coefficients, up to 48K samples. The top 24K samples reduce performance by only 0.51\%, while the top 5K (about 10\% of the context limit) incur only a 1.80\% reduction, showing that much of the predictive signal can be retained with substantially smaller contexts without rerunning optimization.

\textbf{Interpreting VIP Rankings - A Case Study:~} The VIP ranking provides an interpretable view of which samples and features contribute to TFM predictions. To illustrate this capability, we conduct a feature-selection case study on the UCI Communities and Crime dataset containing features for predicting violent crime. After converting the continuous target into a balanced binary classification task, we run \method with sample selection disabled and optimize feature selection for 100 rounds using TabICLv2. The resulting ranking closely agrees with the dataset creator's univariate Pearson-correlation analysis: the top four VIP features exactly match the four features with the largest absolute correlations, and 9 of the creator's top 10 univariate predictors appear in the VIP-COp top 20. VIP also distinguishes among correlated features, capturing context-dependent contributions beyond marginal correlation. Full results are in Appdx.~\ref{app:case_study}.

\begin{figure*}[]
\vspace{-0.1in}
\centering

\begin{minipage}[t]{0.52\textwidth}
\vspace{0pt} 
\centering

\begin{subfigure}[t]{0.48\linewidth}
\includegraphics[width=\linewidth]{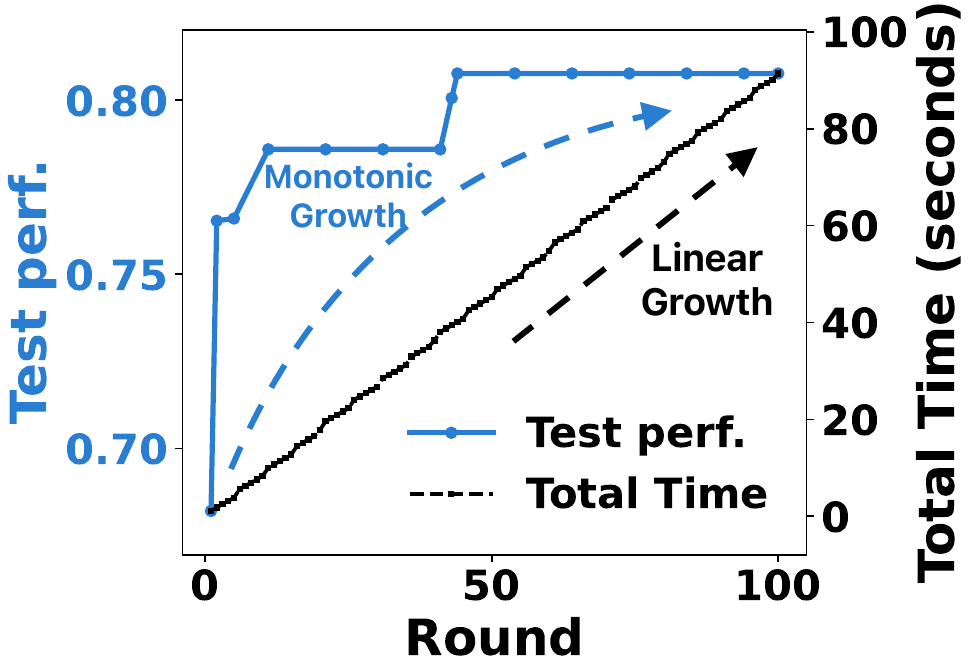}
\caption{RELATHE\_1}
\end{subfigure}
\hfill
\begin{subfigure}[t]{0.48\linewidth}
\includegraphics[width=\linewidth]{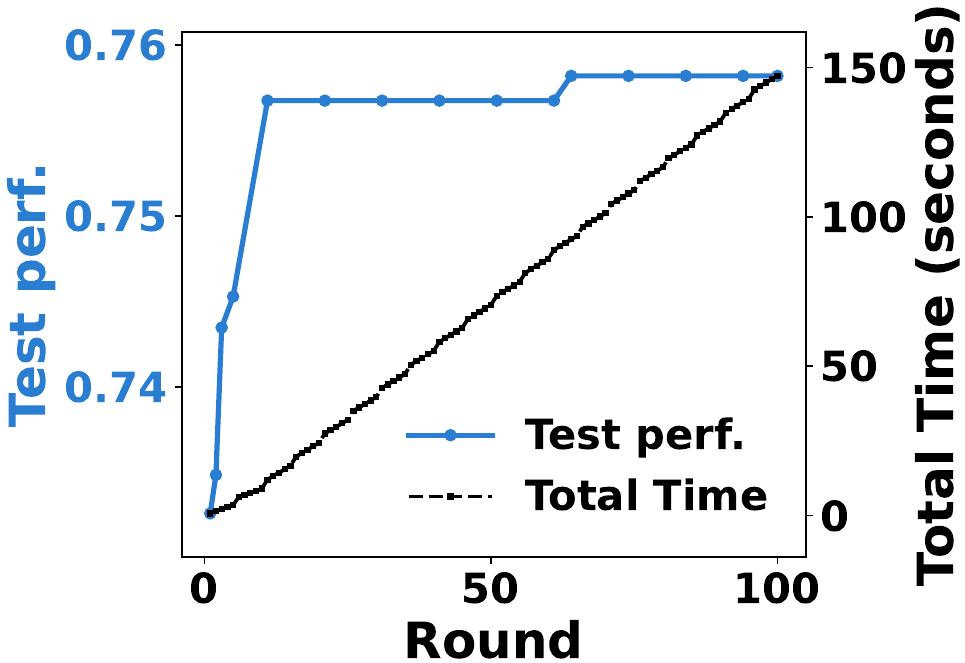}
\caption{jungle\_chess\_endgame}

\end{subfigure}
\begin{subfigure}[t]{0.48\linewidth}
\includegraphics[width=\linewidth]{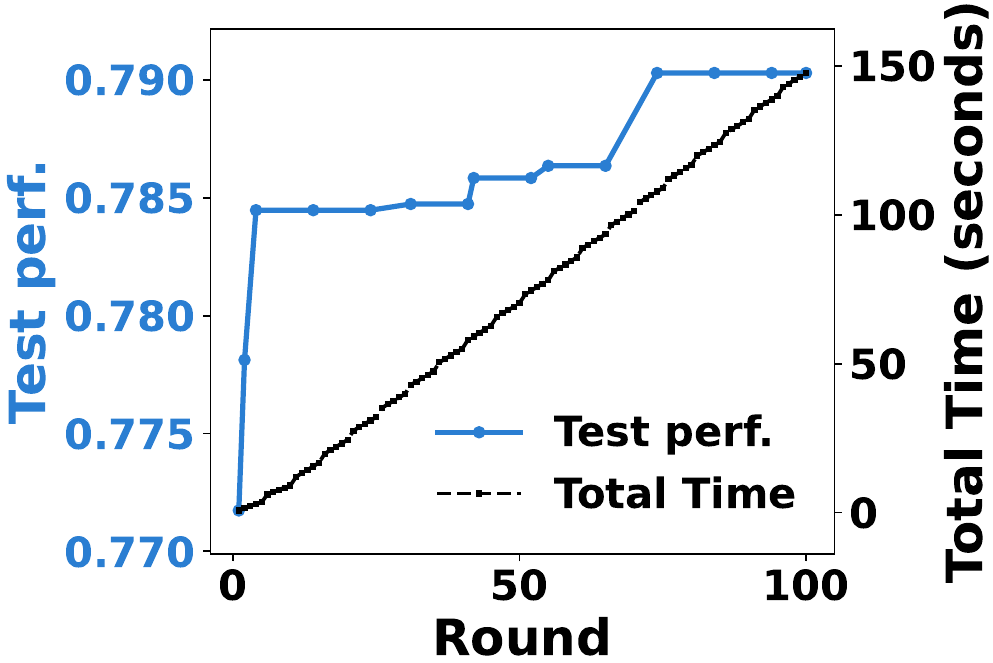}
\caption{INNHotelsGroup}
\end{subfigure}
\hfill
\begin{subfigure}[t]{0.48\linewidth}
\includegraphics[width=\linewidth]{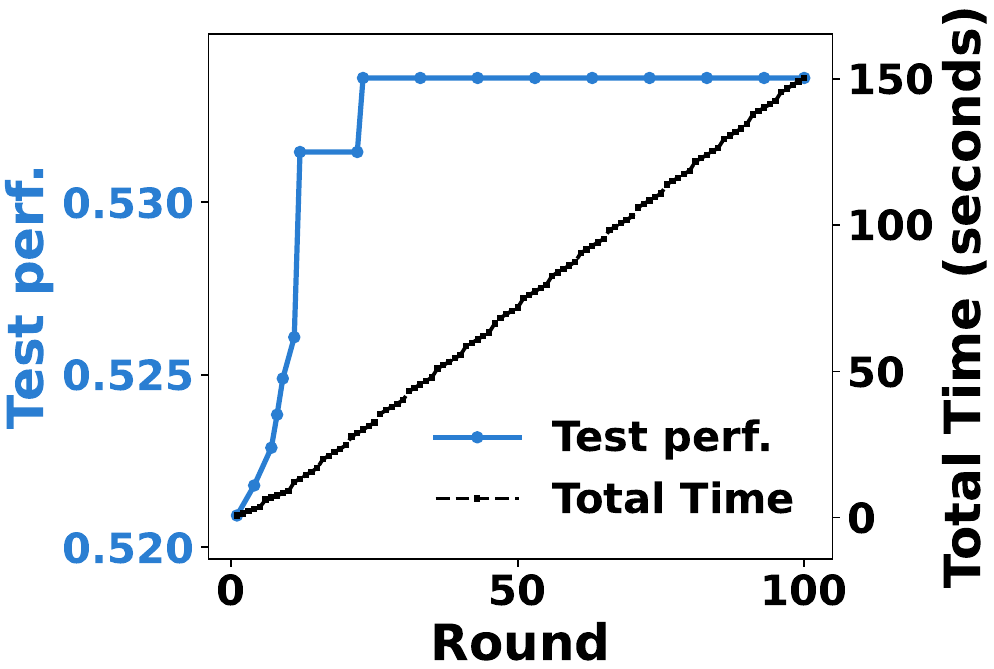}
\caption{BNG(cmc)}

\end{subfigure}
\end{minipage}
\hspace{0.1in}
\raisebox{-0.15in}{%
\begin{minipage}[t]{0.42\textwidth}
\vspace{0pt} 
\centering
\leavevmode\vspace{0.15in} 
\includegraphics[width=0.875\linewidth]{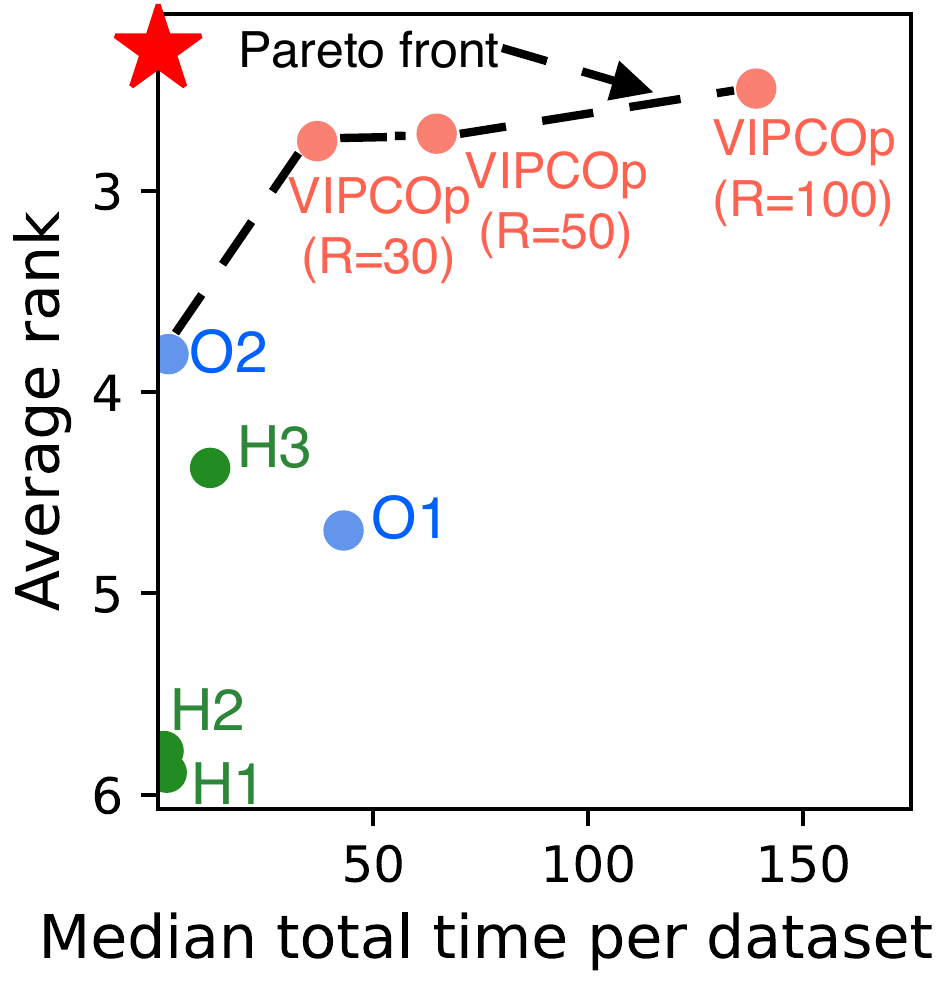}
\end{minipage}
}
\vspace{-0.075in}
\caption{(left) \method \textbf{running time} (black) \textit{increases linearly} with optimization rounds ($\approx$$1.4$s per round), while \textbf{test performance} (blue) \textit{improves monotonically}. (right) \method achieves the best \textbf{performance-time trade-off}, where its anytime variants occupy the Pareto frontier.}
\label{fig:time}
\vspace{-0.05in}
\end{figure*}

\textbf{Running Time, Optimization Efficiency, and Performance:~} Context optimization adds test-time compute and latency, therefore it is important for \method to provide useful gains within a practical budget. Fig.~\ref{fig:time} (left) analyzes \method's performance over increasing time budget. Its running time increases \textit{linearly} with optimization rounds, averaging approximately 1.4, 2.5, and 1.7 seconds per round for TabPFN v1, TabICLv2, and Outformer, respectively, while test performance \textit{improves monotonically}. Thus, \method is an \textit{anytime} procedure that can trade optimization budget for performance. For TabICLv2, 100 rounds take approximately 4 minutes, while 58 of 64 dataset/setting combinations reach their best observed context by round 75, including 22 within the first 25 rounds (see Appdx. Table~\ref{tab:saturation} and  Fig.~\ref{fig:timegrid}). Finally, Fig. \ref{fig:time} (right) illustrates the performance–time trade-off for all methods, showing average rank (lower is better) versus median running time (lower is better) across all datasets. \method and its anytime variants, respectively at rounds 30, 50 and 100, achieve the best performance-runtime trade-off, effectively defining the Pareto frontier.



\section{Conclusion}
Tabular foundation models (TFMs) offer zero-shot classification, regression, and outlier detection in a single forward pass;
however, they trade dataset-specific training for careful context design. 
We address this central challenge with \method, a principled and scalable approach that identifies the most influential samples and features via black-box, model-aware value attribution. Across diverse settings, \method consistently distills oversized, noisy, or uninformative candidate pools into compact, high-signal contexts, yielding robust performance gains under practical test-time budgets. Its anytime nature further enables flexible trade-offs between accuracy and latency. By unifying effectiveness, efficiency, and robustness, \method elevates context optimization from heuristic practice to a core component of TFMs, enabling reliable in-context learning on structured data.

\section*{AI use statement}
In this work, we used generative AI tools to aid and polish writing and to draft sections of the paper. We did not use generative AI tools for retrieval or discovery, research ideation, experimental design or execution, data generation, or generating synthetic datasets. All AI-assisted content was reviewed and verified by the authors. In particular, AI-assisted text was checked for factual accuracy, consistency, and appropriate attribution, and any AI-assisted code was manually reviewed and tested by the authors for correctness before use. The authors take full responsibility for the final content of this work, including all claims, code, and other artifacts produced with the aid of generative AI.

\section*{Reproducibility statement}
We provide the complete algorithmic details of \method, including the optimization procedure, VIP estimation, temperature-guided importance sampling, and context selection, at this link: \url{https://anonymous.4open.science/r/VIP-COp}.
\clearpage

\bibliography{iclr2027_conference}
\bibliographystyle{iclr2027_conference}

\appendix
\section*{Appendix}

\section{Experiment Setup Details}
\label{app:setup}

\subsection{Classification Datasets}
\label{app:datasets}

We sourced classification datasets from the {\sc{TALENT}} and {\sc{TALENT-Extension}} benchmarks \cite{ye2025closer}. For \emph{small-d} datasets ($d \leq 100$), we included all publicly available datasets within the benchmark whose training size exceeded 1,000 samples. For high-dimensional datasets ($d > 100$), we included all publicly available datasets with more than 1,000 samples. We note that {\sc{TALENT-Extension many-class}} benchmark is excluded as the corresponding data sources were not publicly accessible at the time of experimentation Details in Table \ref{tab:datasets}.
\begin{table}[h]
\centering
\caption{Dataset Summary Statistics; categorized by $\Dtr$ sample size $n$ and dimensions $d$ (count in parentheses). ID refers to the identifier within the TALENT benchmarks \cite{ye2025closer}.}
\vspace{0.025in}
\label{tab:datasets}
\begin{tabular}{lllll}
\hline
\textbf{ID} & \textbf{Name (Source)} & \textbf{\#classes} & \textbf{$n$} & \textbf{$d$} \\ \hline
\multicolumn{5}{c}{\textit{small-d} (16)} \\ \toprule
19 & Amazon\_employee\_access & 2 & 26,215 & 7 \\ 
145 & INNHotelsGroup & 2 & 29,020 & 17 \\ 
41 & BNG(breast-w) & 2 & 31,492 & 9 \\ 
47 & BNG(tic-tac-toe) & 2 & 31,492 & 9 \\ 
55 & Click\_prediction & 2 & 31,958 & 3 \\ 
109 & FOREX\_audcad-hour-High & 2 & 35,060 & 10 \\ 
112 & FOREX\_audjpy-hour-High & 2 & 35,060 & 10 \\ 
113 & FOREX\_audsgd-hour-High & 2 & 35,060 & 10 \\ 
114 & FOREX\_audusd-hour-High & 2 & 35,060 & 10 \\ 
116 & FOREX\_cadjpy-hour-High & 2 & 35,060 & 10 \\ 
155 & jungle\_chess\_endgame & 3 & 35,855 & 6 \\ 
30 & bank & 2 & 36,168 & 16 \\ 
93 & electricity & 2 & 36,249 & 8 \\ 
188 & mobile\_c36\_oversampling & 2 & 41,408 & 6 \\ 
42 & BNG(cmc) & 3 & 44,236 & 9 \\ 
248 & shuttle & 7 & 46,400 & 9 \\ \hline
\multicolumn{5}{c}{\textit{small-d (n$>$50K)} (8)} \\ \hline
147 & internet\_firewall & 4 & 52,425 & 7 \\ 
52 & Cardiovascular-Disease & 2 & 56,000 & 11 \\ 
77 & Credit\_c & 3 & 80,000 & 22 \\ 
239 & SDSS17 & 3 & 80,000 & 12 \\ 
80 & diabetes\_us\_hospitals & 2 & 81,412 & 20 \\ 
15 & airline\_satisfaction & 2 & 103,904 & 21 \\ 
229 & Rain\_in\_Australia & 3 & 116,368 & 18 \\ 
8 & accelerometer & 4 & 122,403 & 4 \\\hline
\multicolumn{5}{c}{\textit{Large-d} (14)} \\ \hline
15  [VLS] & nomao & 2 & 27,572 & 118 \\ 
121 & gas-drift & 6 & 8,902 & 128 \\ 
20  [VLS] & volkert & 10 & 46,648 & 180 \\ 
177 & mfeat-factors & 10 & 1,600 & 216 \\ 
144 & Indian\_pines & 8 & 7,315 & 220 \\ 
181 & mfeat-pixel & 10 & 1,600 & 240 \\ 
171 & madeline & 2 & 2,512 & 259 \\ 
8   [VLS] & fabert & 7 & 6,589 & 800 \\ 
10  [VLS] & gina\_agnostic & 2 & 2,774 & 970 \\ 
7   [VLS] & dilbert & 5 & 8,000 & 2,000 \\ 
12  [HD] & PCMAC\_1 & 2 & 1,554 & 3,289 \\ 
14  [HD] & RELATHE\_1 & 2 & 1,141 & 4,322 \\ 
3   [HD] & BASEHOCK\_1 & 2 & 1,594 & 4,862 \\ 
6 & gisette\_1 & 2 & 5,600 & 5,000 \\  
\bottomrule
\end{tabular}
\end{table}

\subsection{Outlier Detection Datasets}
\label{app:odd_datasets}
We sourced all outlier detection datasets from the {\sc{MacrOData}} OddBench \cite{Ding_2026} that are publicly available whose training size exceeded 6,000 samples. 
\setlength{\tabcolsep}{3pt}
\renewcommand{\arraystretch}{0.92}
\small

\begin{longtable}{p{0.30\textwidth}cccccc}
\caption{Dataset Summary Statistics;  Datasets in Alphabetical order \cite{Ding_2026}}
\label{tab:odd_dataset_statistics}\\
\toprule
\textbf{Name (Source)} & \textbf{$n_{\mathrm{rows}}$} & \textbf{$n_{\mathrm{features}}$} &
\textbf{Name (Source)} & \textbf{$n_{\mathrm{rows}}$} & \textbf{$n_{\mathrm{features}}$} \\
\midrule
\endfirsthead

\multicolumn{6}{c}{\tablename\ \thetable{} -- continued from previous page} \\
\toprule
\textbf{Name (Source)} & \textbf{$n_{\mathrm{rows}}$} & \textbf{$n_{\mathrm{features}}$} &
\textbf{Name (Source)} & \textbf{$n_{\mathrm{rows}}$} & \textbf{$n_{\mathrm{features}}$} \\
\midrule
\endhead

\midrule
\multicolumn{6}{r}{Continued on next page} \\
\endfoot

\bottomrule
\endlastfoot

AbandonedMineLand & 13183 & 5 &
AdPerformance & 20426 & 3 \\
AdRequestFraud & 25150 & 77 &
AgilePriority & 17166 & 5 \\
AgriculturalYield & 39448 & 7 &
AircraftSensorAnomalies & 173935 & 19 \\
AmbulanceDeployment & 133267 & 7 &
AnomalyClassification & 262144 & 4 \\
ArtCollections & 19282 & 3 &
ArtSalesPlatform & 46867 & 4 \\
ArxivProcessing & 93811 & 3 &
AstronomicalObservations & 136625 & 11 \\
AutoencoderAnomaly & 166776 & 6 &
BackupSession & 29498 & 5 \\
BankTransactions & 82163 & 3 &
BankruptcyStatus & 78682 & 19 \\
BarExamPrediction & 19178 & 10 &
BaseballEvents & 145585 & 21 \\
BearingFault & 25296 & 10 &
BettingIntegrity & 16309 & 5 \\
BikeSharingTemperature & 17379 & 6 &
BioSequenceAnomaly & 24814 & 5 \\
BiologicalClassifications & 26031 & 13 &
BirdMoult & 19156 & 12 \\
BirdTracking & 17904 & 3 &
BlacklistSpam & 93113 & 6 \\
BlockchainTransaction & 67601 & 3 &
BoliviaUnpaidLabor & 16839 & 64 \\
BondMarketAnomalies & 14412 & 12 &
BuildingCalls & 51010 & 10 \\
BuildingDamage & 15504 & 15 &
BuildingInspection & 14940 & 24 \\
BuildingInspections & 27619 & 5 &
BuildingPoints & 124818 & 8 \\
BusinessFraud & 50000 & 5 &
BusinessInspections & 16103 & 4 \\
BusinessProcessActivity & 118721 & 4 &
CampaignOutcome & 29149 & 7 \\
CancerTreatment & 124394 & 4 &
CandidateVetting & 14575 & 18 \\
CarMakeAnomaly & 25997 & 4 &
CardFraudDetection & 49264 & 4 \\
CardioCatchBMI & 68129 & 7 &
CarrefourSalesDefects & 39058 & 8 \\
CasePriority & 24510 & 27 &
ChecksumValidation & 141804 & 3 \\
ChicagoFoodInspection & 46030 & 11 &
ChronicDiseaseStats & 20349 & 10 \\
CodeDefects & 19337 & 41 &
CogLoadReaction & 76612 & 11 \\
ContactTypeMonitoring & 234512 & 5 &
ConversationAnomalies & 82766 & 6 \\
CoralDisease & 213853 & 8 &
CountyHealthMetrics & 180583 & 4 \\
CourseEvaluation & 419829 & 4 &
CourseRegistration & 25435 & 5 \\
CovidEpidemiology & 16926 & 4 &
CovidTestingData & 15315 & 5 \\
CovtypeAnomaly & 286048 & 10 &
CreditCardFraud & 59367 & 5 \\
CreditDefault & 29019 & 14 &
CreditRiskModel & 145741 & 8 \\
CrimeAnomalies & 94383 & 5 &
CrimeClearanceStatus & 60748 & 13 \\
CrimeFraudDetection & 13307 & 10 &
CropProduction & 33275 & 9 \\
CustomerCampaignOutcome & 21161 & 6 &
CustomerOrderPriority & 33349 & 8 \\
CustomerServiceTransactions & 88186 & 4 &
CybersecurityVulnerability & 21045 & 6 \\
DallasBankGeography & 62964 & 5 &
DamageFunctionSources & 32580 & 3 \\
DataValidationEntries & 38876 & 70 &
DeathCauses & 27622 & 32 \\
DeliveryRootCause & 79464 & 4 &
DentalConditionAnomaly & 32046 & 14 \\
DiabetesSpecialty & 48819 & 14 &
DiamondClarity & 18002 & 10 \\
DifferentialExpression & 104014 & 6 &
DisasterIncidentType & 56981 & 4 \\
DoctorVisits & 23713 & 7 &
DragonFlySystem & 22109 & 4 \\
DurhamVotersAge & 28379 & 4 &
EcommerceOrderPriority & 33359 & 7 \\
ElectricBuildingUsage & 312958 & 10 &
ElectronicTesting & 75387 & 12 \\
EmailPhishingDetection & 45793 & 16 &
EmotionalIncidents & 36614 & 6 \\
EmployeeAttrition & 51350 & 11 &
EmployeeLocationAnomalies & 333952 & 3 \\
EndangeredSpeciesClassification & 173476 & 3 &
GPUResourceUsage & 21886 & 4 \\
GeographicSeverity & 122604 & 3 &
GeologicalLithology & 225455 & 4 \\
GeospatialCrimeResolution & 129957 & 10 &
GlobalSharkAttacks & 19974 & 3 \\
GlobalTerrorismAttack & 33620 & 5 &
HealthMetricsAnomalies & 96476 & 8 \\
HealthPulse & 44119 & 28 &
HealthcareMessages & 24378 & 8 \\
HockeyGameplay & 125494 & 5 &
HospitalSeverity & 109369 & 9 \\
HousePriority & 16849 & 3 &
HousingConstruction & 21424 & 9 \\
HuntingCriteria & 14486 & 5 &
ImageCategoryAnomaly & 82273 & 6 \\
IndustrialEquipment & 29418 & 10 &
IndustrialFanVibration & 440178 & 6 \\
ItemIncineration & 23080 & 14 &
JobStatus & 38081 & 19 \\
JobStatusMonitoring & 32696 & 4 &
KidneyDisease & 42941 & 9 \\
LabTestAbnormality & 18225 & 7 &
LandParcels & 35865 & 3 \\
LandShapeTrade & 73481 & 8 &
LanguageAnomaly & 17760 & 14 \\
LanguageLocationAnomaly & 368218 & 3 &
LargeWoodMeasurement & 54331 & 9 \\
LaunchStatus & 54573 & 3 &
LegalChargeAnomalies & 23571 & 5 \\
LegalEducation & 27472 & 6 &
LoanCreditScore & 102617 & 5 \\
LoanFraudDetection & 20269 & 9 &
LoyaltyFraud & 100000 & 7 \\
MaritimeVessel & 229235 & 8 &
MaterialFlow & 382500 & 9 \\
MechanicalTesting & 28014 & 6 &
MedicalClassificationAnomalies & 31103 & 4 \\
MicrostructureDefect & 32712 & 18 &
MineralComposition & 27347 & 17 \\
MobileMoneyTransactions & 15005 & 7 &
MojaveDune & 14760 & 17 \\
MouseInteractionPhase & 29155 & 5 &
MovieGenreAnomalies & 20007 & 4 \\
MusicFamiliarity & 22025 & 5 &
NestSuccess & 30947 & 12 \\
NetworkMalware & 40837 & 15 &
NetworkResponseAnomalies & 14413 & 9 \\
NetworkThreat & 12993 & 4 &
NetworkTrafficDOS & 197342 & 4 \\
NetworkTrafficSpam & 181803 & 4 &
NewbornReferral & 15488 & 3 \\
NonComplianceInspection & 57410 & 5 &
OceanicStatus & 37752 & 20 \\
OptimizationError & 25481 & 17 &
OrderTransmission & 31864 & 3 \\
OzoneLevels & 342858 & 6 &
PatientReadmission & 78887 & 11 \\
PhotoAnomalyDetection & 12952 & 3 &
PlayerMatchEvents & 335787 & 9 \\
PokemonTypeAnomaly & 117184 & 8 &
PoliceFraudIncidents & 508947 & 4 \\
PoliticalInterestMisinformation & 23680 & 9 &
PopulationComparison & 80035 & 6 \\
PostcodeValidation & 15435 & 9 &
PrisonBooking & 38408 & 5 \\
ProcessState & 18581 & 5 &
ProductionRunStatus & 43992 & 7 \\
ProgrammingAnomalies & 32618 & 6 &
ProjectSeverity & 16548 & 4 \\
PropertyCondition & 24649 & 7 &
ProteinCodonAnalysis & 32057 & 14 \\
ProtestEvents & 24366 & 4 &
ProtestViolence & 20224 & 7 \\
PsychTrialAccuracy & 55738 & 5 &
PsychologicalTrialValidity & 16128 & 5 \\
PsychologyExperiment & 23562 & 11 &
RealEstateCondition & 87400 & 5 \\
RealEstateProperty & 18599 & 25 &
RegionalPriceMonitoring & 20943 & 6 \\
ReportResolution & 87154 & 3 &
ResidueLigandClassification & 14498 & 9 \\
RestaurantViolation & 12756 & 6 &
RiskVulnerability & 35154 & 3 \\
RoutingAnomaly & 66473 & 6 &
SalesOrderPriority & 33386 & 7 \\
SalesTaskStatus & 14796 & 15 &
SatallaxBenchmark & 20948 & 4 \\
SaudiWeatherAnomalies & 244536 & 9 &
SchedulerJobState & 57595 & 7 \\
SensorimotorAdaption & 23322 & 19 &
SentencingFraud & 15174 & 53 \\
ServerLoadTest & 21831 & 3 &
ServiceErrorLocation & 16215 & 4 \\
ShapeTrackingAnomaly & 115129 & 6 &
ShellfishQuality & 100000 & 5 \\
ShuttleOutlier & 41638 & 9 &
SierraLeoneWaterPoints & 24173 & 6 \\
SmallBusinessLoans & 29151 & 12 &
SnowDepthMonitoring & 32986 & 3 \\
SoccerActions & 45566 & 25 &
SoccerMatchAnomalies & 17792 & 8 \\
SoilFloodProperties & 20301 & 4 &
SpamDetection & 67881 & 5 \\
SpeciesObservation & 49688 & 8 &
SpecimenStatus & 27296 & 6 \\
StatisticalTestResults & 42601 & 10 &
StoreCrowd & 844338 & 3 \\
StormBehavior & 18512 & 6 &
StudentAssessment & 117017 & 10 \\
StudentEnrollment & 56106 & 4 &
StudentProgramAnomaly & 169579 & 3 \\
SummaryAnomalies & 35050 & 3 &
SuperStoreOrderPriority & 33339 & 10 \\
SystemFailure & 28719 & 11 &
SystemFailureMonitoring & 25010 & 6 \\
SystemLogs & 21834 & 3 &
SystemProcessAnomalies & 23121 & 3 \\
SystemStatusAnomaly & 19789 & 5 &
TechnicalAnomalies & 20020 & 8 \\
TelecomMarketing & 37471 & 7 &
TestCycleAnomalies & 56763 & 3 \\
TVShowAnomalies & 107613 & 3 &
TicketPriority & 30333 & 7 \\
TorontoPoliceContacts & 25802 & 5 &
TrafficIncident & 56429 & 5 \\
TrafficStopAnomalies & 49787 & 4 &
TransactionFeeAnomalies & 44688 & 3 \\
TransactionService & 18311 & 5 &
TrialConditionAnomaly & 126330 & 3 \\
TurnaroundValidation & 34615 & 3 &
TwitchStreamAnomalies & 131024 & 20 \\
UserEventAnomalies & 20738 & 3 &
VehicleNetworkTraffic & 24895 & 59 \\
VenezuelaBeneficiaries & 16841 & 6 &
VideoGameViolence & 19627 & 3 \\
VolcanicAnomalies & 27475 & 4 &
VolleyballMatch & 26305 & 38 \\
VoteOutcome & 26898 & 8 &
WaterMeterAnomaly & 19812 & 5 \\
WaterPointFunctionality & 26076 & 13 &
WaterQualityAssessment & 121722 & 8 \\
WeatherEvents & 25533 & 7 &
WikipediaVandalism & 29646 & 3 \\
WindTurbineFault & 47389 & 4 &
WindmillAnomaly & 17220 & 17 \\
WindowsEditionAnomaly & 49384 & 26 &
WorkOvertimePatterns & 18151 & 5 \\
YouTubeFaceDetection & 55036 & 5 &
& & \\

\end{longtable}

\clearpage
\subsection{Baseline Configurations}
\label{app:baseconfigs}

All baseline experiments are implemented through a unifed TabPFN wrapper interface and performance is evaluated using balanced accuracy. The global seed is fixed to 42 for all operations that involve stochasticity unless otherwise stated. Each method receives the same train/validation/test split and uses identical TabPFN-v1 backbone model. For experiments involving data augmentation (DA) or data noise (DN), the processed datasets and their associated splits are generated once and stored, which are then reused across all baseline methods to ensure an identical evaluation setup. During inference, test examples are evaluated in batches of 2000 samples per forward pass. The max context configuration, denoted (\emph{MAX\_CONTEXT}, \emph{MAX\_FEATURES}) is set to (\emph{1000}, \emph{100}) for TabPFN-v1.

\textbf{H1: Random}. H1 performs 15 independent random context runs. If $n >$ \emph{MAX\_CONTEXT}, each run samples exactly \emph{MAX\_CONTEXT} training examples uniformly without replacement; otherwise, all training examples are used towards context. If $d >$ \emph{MAX\_FEATURES}, a size \emph{MAX\_FEATURES} feature subset is sampled once per run using the fixed seed. The final result is the average of test balanced accuracy across the 15 runs.

\textbf{H2: Ensemble}. H2 performs 20 runs and assembles the predictions. For each run, if $n >$ \emph{MAX\_CONTEXT}, one random sample subset of size \emph{MAX\_CONTEXT} is selected. Similarly, one random feature subset of size \emph{MAX\_FEATURES} is selected if $d >$ \emph{MAX\_FEATURES}. Predictions from 20 independent samples are averaged at the probability level before computing balanced accuracy.

\textbf{H3: XLContext}. H3 uses the full training set  as context, i.e., without applying \emph{MAX\_CONTEXT} sample restriction. However, if $d >$ \emph{MAX\_FEATURES}, features are randomly subsampled to \emph{MAX\_FEATURES} using the fixed seed. If the inference triggers a CUDA out-of-memory error (depending on GPU memory), the context size is multiplied by 0.9. The method then samples a random subset of the reduced size and retries inference. This process repeats until inference succeeds.

\textbf{O1: KMeansReps}. O1 uses the KMeans implementation from \emph{sklearn.cluster} for both feature and sample selection, with feature selection performed before sample selection. The number of K-means initializations is defaulted to 5. For feature-level optimization, if $d >$ \emph{MAX\_FEATURES}, each feature is treated as a point by transposing the training matrix. KMeans is then run with $n\_clusters =$ \emph{MAX\_FEATURES}. For each centroid, the nearest unique feature is selected. Each initialization is evaluated on the validation set using balanced accuracy, and the best feature subset is retained. Then, if $n >$ \emph{MAX\_CONTEXT}, Kmeans is run on the reduced training matrix with $n\_clusters =$ \emph{MAX\_CONTEXT}. The nearest unique training example to each centroid is selected. Validation balanced accuracy is used to choose the best initialization. The selected feature indices and context indices are then fixed for final test evaluation.

\textbf{O2: DT+TFM}.  O2 implementation uses a decision tree for sample routing. If $d >$ \emph{MAX\_FEATURES}, features are first selected uniformly at random without replacement using the fixed seed. Then, if $n >$ \emph{MAX\_CONTEXT}, a DecisionTreeClassifier is fitted on the reduced training matrix with gini impurity, $min\_samples\_leaf =$ \emph{MAX\_CONTEXT}, and $random\_state =$ seed. During evaluation, each test example is routed to a leaf, and the training samples assigned to that leaf are used as its local TabPFN context. If a leaf contains more than \emph{MAX\_CONTEXT} examples, only the first \emph{MAX\_CONTEXT} examples are retained.

\textbf{Hardware Configuration:~} We based our {\sc{VIP-COp}\xspace} and baseline experiments on 4 NVIDIA Tesla V100-SXM2-32GB GPUs with 2 Intel Xeon Gold 6248 20-Core Processors.

\clearpage

\section{Detailed Experiment Results}
\label{app:results}

\begin{table}[!th]
\centering
\caption{\textbf{TabPFNv1} Performance results using each method on individual datasets under the \textbf{original HardCOp} setting. Best performance per dataset is shown in \textbf{bold}, with average percentage (\%) improvement by \method against each Baseline is highlighted \colorbox{blue!20}{in color}.}
\label{tab:og_results_tab}
\resizebox{\textwidth}{!}{
\begin{tabular}{llcccccc}

\toprule
 & \textbf{Name} & \textbf{VIP\_COp} & \textbf{H1} & \textbf{H2} & \textbf{H3} & \textbf{O1} & \textbf{O2} \\

\textbf{ID} & \textbf{(Source)} & \textbf{(ours)} & \textbf{(Mean15)} & \textbf{(Ensem20)} & \textbf{(XXL)} & \textbf{(Kmeans)} & \textbf{(DT+TFM)} \\ \midrule 

\multicolumn{8}{c}{\textit{small-d (16)}} \\ \hline
19 & Amazon\_employee\_access & \textbf{0.500} & \textbf{0.500} & \textbf{0.500} & \textbf{0.500} & \textbf{0.500} & \textbf{0.500}  \\
145 & INNHotelsGroup & 0.790 & 0.775 & 0.788 & 0.793 & 0.790 & \textbf{0.820}  \\
41 & BNG(breast-w) & 0.978 & 0.978 & 0.980 & \textbf{0.983} & 0.980 & 0.982  \\
47 & BNG(tic-tac-toe) & 0.704 & 0.691 & 0.683 & 0.715 & 0.713 & \textbf{0.757}  \\
55 & Click\_prediction & \textbf{0.504} & 0.500 & 0.500 & 0.501 & 0.500 & 0.500  \\
109 & FOREX\_audcad-hour-High & 0.676 & 0.611 & 0.630 & \textbf{0.679} & 0.606 & 0.667  \\
112 & FOREX\_audjpy-hour-High & \textbf{0.618} & 0.511 & 0.500 & \textbf{0.618} & 0.591 & 0.534  \\
113 & FOREX\_audsgd-hour-High & \textbf{0.537} & 0.501 & 0.500 & 0.500 & 0.520 & 0.522  \\
114 & FOREX\_audusd-hour-High & \textbf{0.520} & 0.500 & 0.500 & 0.500 & 0.500 & 0.502  \\
116 & FOREX\_cadjpy-hour-High & 0.573 & 0.504 & \textbf{0.598} & 0.500 & 0.527 & 0.515  \\
155 & jungle\_chess\_endgame & 0.758 & 0.662 & 0.670 & 0.669 & 0.718 & \textbf{0.793}  \\
30 & bank & \textbf{0.752} & 0.619 & 0.588 & 0.619 & 0.644 & 0.634  \\
93 & electricity & 0.765 & 0.759 & 0.759 & 0.757 & 0.764 & \textbf{0.817}  \\
188 & mobile\_c36\_oversampling & 0.904 & 0.906 & 0.909 & 0.914 & 0.909 & \textbf{0.955}  \\
42 & BNG(cmc) & 0.534 & 0.523 & 0.518 & 0.542 & \textbf{0.545} & 0.529  \\
248 & shuttle & 0.673 & 0.555 & 0.550 & 0.729 & \textbf{0.929} & 0.619  \\

\multicolumn{2}{l}{\textit{Avg. \% Improv. over Baseline}} & -- & \colorbox{cyan!20}{\textbf{7.56\%}} & \colorbox{cyan!20}{\textbf{6.80\%}} & \colorbox{cyan!20}{\textbf{3.08\%}} & \colorbox{cyan!20}{\textbf{1.49\%}} & \colorbox{cyan!20}{\textbf{2.29\%}}  \\
\hline

\multicolumn{8}{c}{\textit{small-d (n$>$50K) (8)}} \\ \hline
147 & internet\_firewall & \textbf{0.498} & \textbf{0.498} & \textbf{0.498} & 0.497 & 0.250 & \textbf{0.498}  \\
52 & Cardiovascular-Disease & 0.705 & 0.709 & \textbf{0.721} & 0.709 & 0.674 & \textbf{0.721}  \\
77 & Credit\_c & 0.669 & 0.591 & 0.580 & 0.608 & 0.638 & \textbf{0.685}  \\
239 & SDSS17 & 0.956 & 0.950 & 0.958 & \textbf{0.961} & 0.943 & 0.956  \\
80 & dabetes\_us\_hospitals & 0.596 & 0.576 & 0.592 & 0.594 & 0.594 & \textbf{0.597}  \\
15 & airline\_satisfaction & 0.886 & 0.890 & 0.893 & 0.925 & 0.892 & \textbf{0.940}  \\
229 & Rain\_in\_Australia & \textbf{0.688} & 0.550 & 0.484 & 0.635 & 0.660 & 0.652  \\
8 & accelerometer & 0.699 & 0.440 & 0.241 & 0.447 & 0.436 & \textbf{0.729}  \\

\multicolumn{2}{l}{\textit{Avg. \% Improv. over Baseline}} & -- & \colorbox{cyan!20}{\textbf{12.53\%}} & \colorbox{cyan!20}{\textbf{30.62\%}} & \colorbox{cyan!20}{\textbf{8.75\%}} & \colorbox{cyan!20}{\textbf{21.78\%}} & \colorbox{cyan!20}{\textbf{-1.13\%}}  \\
\hline

\multicolumn{8}{c}{\textit{Large-d (14)}} \\ \hline
15 [VLS] & nomao & 0.934 & 0.919 & 0.928 & \textbf{0.944} & 0.932 & 0.938  \\
121 & gas-drift & 0.980 & 0.966 & 0.985 & 0.983 & 0.973 & \textbf{0.986}  \\
20 [VLS] & volkert & 0.339 & 0.296 & 0.295 & 0.330 & 0.321 & \textbf{0.391}  \\
177 & mfeat-factors & 0.950 & 0.953 & \textbf{0.970} & 0.958 & 0.955 & 0.958  \\
144 & Indian\_pines & 0.765 & 0.685 & 0.717 & 0.762 & 0.709 & \textbf{0.803}  \\
181 & mfeat-pixel & 0.933 & 0.933 & \textbf{0.948} & 0.938 & 0.938 & 0.938  \\
171 & madeline & 0.610 & 0.581 & 0.540 & 0.614 & 0.608 & \textbf{0.637}  \\
8 [VLS] & fabert & 0.275 & 0.234 & 0.217 & 0.261 & \textbf{0.281} & 0.250  \\
10 [VLS] & gina\_agnostic & 0.876 & 0.802 & 0.836 & 0.814 & 0.843 & \textbf{0.882}  \\
7 [VLS] & dilbert & 0.881 & 0.831 & 0.830 & \textbf{0.927} & 0.904 & \textbf{0.927}  \\
12 [HD] & PCMAC\_1 & \textbf{0.865} & 0.671 & 0.578 & 0.670 & 0.717 & 0.703  \\
14 [HD] & RELATHE\_1 & \textbf{0.825} & 0.579 & 0.569 & 0.579 & 0.676 & 0.707  \\
3 [HD] & BASEHOCK\_1 & \textbf{0.890} & 0.658 & 0.657 & 0.677 & 0.670 & 0.815  \\
6 & gisette\_1 & 0.944 & 0.844 & 0.899 & 0.871 & \textbf{0.947} & 0.926  \\

\multicolumn{2}{l}{\textit{Avg. \% Improv. over Baseline}} & -- & \colorbox{cyan!20}{\textbf{13.23\%}} & \colorbox{cyan!20}{\textbf{14.56\%}} & \colorbox{cyan!20}{\textbf{8.51\%}} & \colorbox{cyan!20}{\textbf{6.30\%}} & \colorbox{cyan!20}{\textbf{2.18\%}}  \\
\midrule
\multicolumn{2}{l}{\textbf{\textit{Overall Avg. \% Improv. over Baseline}}} & -- & \colorbox{blue!20}{\textbf{10.69\%}} & \colorbox{blue!20}{\textbf{14.68\%}} & \colorbox{blue!20}{\textbf{6.26\%}} & \colorbox{blue!20}{\textbf{7.53\%}} & \colorbox{blue!20}{\textbf{1.54\%}}  \\

\bottomrule

\end{tabular}
}
\end{table}

\begin{figure}[h]
\vspace{-0.1in}
\centering
\includegraphics[width=0.7\linewidth]{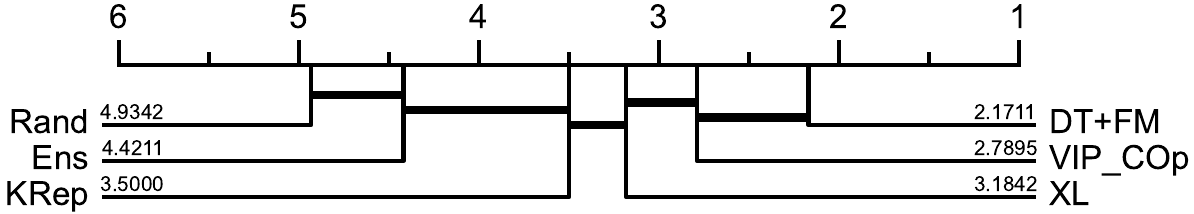}
\caption{CD diagram across datasets (numbers depict average rank) in the original HardCOp setting.}
\vspace{-0.1in}
\label{fig:maindemsar}
\end{figure}

\clearpage
\begin{table}[!th]

\centering

\caption{\textbf{TabICLv2} Performance results using each method on individual datasets under the \textbf{original HardCOp} setting. Best performance per dataset is shown in \textbf{bold}, with average percentage (\%) improvement by \method against each Baseline highlighted \colorbox{blue!20}{in color}.}

\label{tab:TabICL_OG_results_tab}

\resizebox{\textwidth}{!}{

\begin{tabular}{llcccccc}

\toprule

& \textbf{Name} & \textbf{VIP\_COp} & \textbf{H1} & \textbf{H2} & \textbf{H3} & \textbf{O1} & \textbf{O2} \\

\textbf{ID} & \textbf{(Source)} & \textbf{(ours)} & \textbf{(Mean15)} & \textbf{(Ensem20)} & \textbf{(XXL)} & \textbf{(KMeans)} & \textbf{(DT+TFM)} \\

\midrule

\multicolumn{8}{c}{\textit{small-d (16)}} \\ \hline

19 & Amazon\_employee\_access
& \textbf{0.595} & 0.585 & 0.585 & 0.585 & 0.585 & 0.585 \\

145 & INNHotelsGroup
& 0.874 & \textbf{0.882} & \textbf{0.882} & \textbf{0.882} & \textbf{0.882} & \textbf{0.882} \\

41 & BNG(breast-w)
& 0.987 & \textbf{0.988} & \textbf{0.988} & \textbf{0.988} & \textbf{0.988} & \textbf{0.988} \\

47 & BNG(tic-tac-toe)
& \textbf{0.789} & 0.784 & 0.784 & 0.784 & 0.784 & 0.784 \\

55 & Click\_prediction
& \textbf{0.578} & 0.521 & 0.521 & 0.521 & 0.521 & 0.521 \\

109 & FOREX\_audcad-hour-High
& \textbf{0.714} & 0.703 & 0.703 & 0.703 & 0.703 & 0.703 \\

112 & FOREX\_audjpy-hour-High
& \textbf{0.721} & 0.716 & 0.716 & 0.716 & 0.716 & 0.716 \\

113 & FOREX\_audsgd-hour-High
& \textbf{0.707} & 0.656 & 0.656 & 0.656 & 0.656 & 0.656 \\

114 & FOREX\_audusd-hour-High
& \textbf{0.706} & 0.705 & 0.705 & 0.705 & 0.705 & 0.705 \\

116 & FOREX\_cadjpy-hour-High
& \textbf{0.716} & 0.712 & 0.712 & 0.712 & 0.712 & 0.712 \\

155 & jungle\_chess\_endgame
& \textbf{0.858} & 0.841 & 0.841 & 0.841 & 0.841 & 0.841 \\

30 & bank
& \textbf{0.827} & 0.757 & 0.757 & 0.757 & 0.757 & 0.757 \\

93 & electricity
& 0.889 & \textbf{0.914} & \textbf{0.914} & \textbf{0.914} & \textbf{0.914} & \textbf{0.914} \\

188 & mobile\_c36\_oversampling
& 0.989 & \textbf{0.992} & \textbf{0.992} & \textbf{0.992} & \textbf{0.992} & \textbf{0.992} \\

42 & BNG(cmc)
& \textbf{0.578} & 0.563 & 0.563 & 0.563 & 0.563 & 0.563 \\

248 & shuttle
& \textbf{1.000} & 0.991 & 0.991 & 0.991 & 0.991 & 0.991 \\

\multicolumn{2}{l}{\textit{Avg. \% Improv. over Baseline}}
& -- 
& \colorbox{cyan!20}{\textbf{2.18\%}}
& \colorbox{cyan!20}{\textbf{2.18\%}}
& \colorbox{cyan!20}{\textbf{2.18\%}}
& \colorbox{cyan!20}{\textbf{2.18\%}}
& \colorbox{cyan!20}{\textbf{2.18\%}} \\

\hline

\multicolumn{8}{c}{\textit{small-d (n$>$50K) (8)}} \\ \hline

147 & internet\_firewall
& \textbf{0.622} & 0.571 & 0.571 & 0.571 & 0.571 & 0.582 \\

52 & Cardiovascular-Disease
& \textbf{0.735} & 0.733 & 0.733 & 0.734 & 0.733 & 0.732 \\

77 & Credit\_c
& \textbf{0.796} & 0.807 & 0.802 & 0.805 & \textbf{0.808} & 0.801 \\

239 & SDSS17
& \textbf{0.973} & 0.972 & 0.972 & 0.972 & 0.972 & 0.971 \\

80 & dabetes\_us\_hospitals
& 0.633 & 0.631 & 0.631 & 0.631 & \textbf{0.633} & 0.630 \\

15 & airline\_satisfaction
& \textbf{0.974} & 0.944 & 0.938 & 0.949 & 0.944 & 0.944 \\

229 & Rain\_in\_Australia
& \textbf{0.705} & 0.679 & 0.681 & 0.679 & 0.680 & 0.680 \\

8 & accelerometer
& 0.804 & 0.802 & \textbf{0.807} & 0.803 & 0.804 & 0.804 \\

\multicolumn{2}{l}{\textit{Avg. \% Improv. over Baseline}}
& --
& \colorbox{cyan!20}{\textbf{1.95\%}}
& \colorbox{cyan!20}{\textbf{1.98\%}}
& \colorbox{cyan!20}{\textbf{1.87\%}}
& \colorbox{cyan!20}{\textbf{1.82\%}}
& \colorbox{cyan!20}{\textbf{1.74\%}} \\

\midrule

\multicolumn{2}{l}{\textbf{\textit{Overall Avg. \% Improv. over Baseline}}}
& --
& \colorbox{blue!20}{\textbf{2.10\%}}
& \colorbox{blue!20}{\textbf{2.11\%}}
& \colorbox{blue!20}{\textbf{2.08\%}}
& \colorbox{blue!20}{\textbf{2.06\%}}
& \colorbox{blue!20}{\textbf{2.03\%}} \\

\bottomrule

\end{tabular}

}

\end{table}
\begin{table}[!th]
\centering

\caption{\textbf{TabICLv2} Performance results using each method on individual datasets under the \textbf{HardCOp4DA} setting with \textbf{sample augmentation} $\text{S}_{\text{aug}}$. Best performance per dataset is shown in \textbf{bold}, with average percentage (\%) improvement by \method against each Baseline is highlighted \colorbox{blue!20}{in color}. Effect of sample augmentation is shown as the average percentage (\%) difference from performance on the original datasets, highlighted \colorbox{teal!40}{in green}.}

\label{tab:v2_aug_small_d}

\resizebox{\textwidth}{!}{%
\begin{tabular}{llcccccc}

\toprule

& \textbf{Name} 
& \textbf{VIP\_COp} 
& \textbf{H1} 
& \textbf{H2} 
& \textbf{H3} 
& \textbf{O1} 
& \textbf{O2} \\

\textbf{ID} 
& \textbf{(Source)} 
& \textbf{(ours)} 
& \textbf{(Mean15)} 
& \textbf{(Ensem20)} 
& \textbf{(XXL)} 
& \textbf{(KMeans)} 
& \textbf{(DT+TFM)} \\

\midrule

\multicolumn{8}{c}{\textit{small-d (16)}} \\ \hline

19 & Amazon\_employee\_access
& \textbf{0.606} & 0.564 & 0.557 & 0.585 & 0.572 & 0.569 \\

145 & INNHotelsGroup
& \textbf{0.888} & 0.865 & 0.874 & 0.881 & 0.868 & 0.872 \\

41 & BNG(breast-w)
& \textbf{0.987} & 0.985 & 0.985 & \textbf{0.987} & 0.979 & 0.984 \\

47 & BNG(tic-tac-toe)
& \textbf{0.797} & 0.772 & 0.775 & 0.779 & 0.776 & 0.780 \\

55 & Click\_prediction
& \textbf{0.596} & 0.514 & 0.514 & 0.512 & 0.509 & 0.512 \\

109 & FOREX\_audcad-hour-High
& \textbf{0.728} & 0.707 & 0.709 & 0.707 & 0.701 & 0.706 \\

112 & FOREX\_audjpy-hour-High
& \textbf{0.737} & 0.720 & 0.721 & 0.720 & 0.716 & 0.722 \\

113 & FOREX\_audsgd-hour-High
& \textbf{0.724} & 0.705 & 0.708 & 0.702 & 0.699 & 0.701 \\

114 & FOREX\_audusd-hour-High
& \textbf{0.721} & 0.706 & 0.710 & 0.705 & 0.701 & 0.707 \\

116 & FOREX\_cadjpy-hour-High
& \textbf{0.731} & 0.713 & 0.715 & 0.712 & 0.707 & 0.714 \\

155 & jungle\_chess\_endgame
& \textbf{0.860} & 0.810 & 0.814 & 0.839 & 0.840 & 0.844 \\

30 & bank
& \textbf{0.898} & 0.718 & 0.721 & 0.736 & 0.728 & 0.728 \\

93 & electricity
& 0.898 & 0.876 & 0.885 & \textbf{0.909} & 0.898 & 0.896 \\

188 & mobile\_c36\_oversampling
& \textbf{0.996} & 0.975 & 0.977 & 0.977 & 0.967 & 0.976 \\

42 & BNG(cmc)
& \textbf{0.588} & 0.565 & 0.568 & 0.567 & 0.564 & 0.570 \\

248 & shuttle
& \textbf{1.000} & 0.903 & 0.916 & 0.987 & 0.950 & 0.916 \\

\multicolumn{2}{l}{\textit{Avg. \% Improv. over Baseline}}
& \textbf{--}
& \colorbox{blue!20}{\textbf{5.80\%}}
& \colorbox{blue!20}{\textbf{5.42\%}}
& \colorbox{blue!20}{\textbf{4.17\%}}
& \colorbox{blue!20}{\textbf{5.23\%}}
& \colorbox{blue!20}{\textbf{5.00\%}} \\

\multicolumn{2}{l}{\textbf{\textit{Overall Avg. \% $\Delta$ from Original (Table \ref{tab:TabICL_OG_results_tab})}}} & \colorbox{teal!40}{\textbf{1.91\%}} & \colorbox{teal!40}{\textbf{-1.43\%}} & \colorbox{teal!40}{\textbf{-1.08\%}} & \colorbox{teal!40}{\textbf{0.05\%}} & \colorbox{teal!40}{\textbf{-0.95\%}} & \colorbox{teal!40}{\textbf{-0.70\%}} \\
\bottomrule

\end{tabular}
}

\end{table}


\begin{table}[!th]
\centering
\caption{\textbf{TabPFNv1} Performance results using each method on individual datasets under the \textbf{HardCOp4DA} setting with \textbf{sample augmentation} $\text{S}_{\text{aug}}$. Best performance per dataset is shown in \textbf{bold}, with average percentage (\%) improvement by \method against each Baseline is highlighted \colorbox{blue!20}{in color}. Effect of sample augmentation is shown as the average percentage (\%) difference from performance on the original datasets, highlighted \colorbox{teal!40}{in green}.}
\label{tab:aug_sample_results_tab}
\resizebox{\textwidth}{!}{
\begin{tabular}{llcccccc}

\toprule
 & \textbf{Name} & \textbf{VIP\_COp} & \textbf{H1} & \textbf{H2} & \textbf{H3} & \textbf{O1} & \textbf{O2} \\

\textbf{ID} & \textbf{(Source)} & \textbf{(ours)} & \textbf{(Mean15)} & \textbf{(Ensem20)} & \textbf{(XXL)} & \textbf{(Kmeans)} & \textbf{(DT+TFM)} \\ \midrule

\multicolumn{8}{c}{\textit{small-d (16)}} \\ \hline
19 & Amazon\_employee\_access & 0.508 & 0.500 & 0.505 & 0.500 & 0.500 & \textbf{0.593}  \\
145 & INNHotelsGroup & 0.805 & 0.788 & 0.790 & 0.812 & 0.792 & \textbf{0.853}  \\
41 & BNG(breast-w) & 0.980 & 0.979 & 0.979 & 0.983 & 0.980 & \textbf{0.984}  \\
47 & BNG(tic-tac-toe) & 0.706 & 0.710 & 0.707 & 0.733 & 0.720 & \textbf{0.749}  \\
55 & Click\_prediction & \textbf{0.582} & 0.500 & 0.500 & 0.509 & 0.507 & 0.512  \\
109 & FOREX\_audcad-hour-High & 0.666 & 0.615 & 0.500 & \textbf{0.677} & 0.601 & 0.598  \\
112 & FOREX\_audjpy-hour-High & 0.608 & 0.526 & 0.500 & \textbf{0.617} & 0.589 & 0.553  \\
113 & FOREX\_audsgd-hour-High & 0.537 & 0.504 & 0.500 & 0.500 & \textbf{0.538} & 0.516  \\
114 & FOREX\_audusd-hour-High & \textbf{0.531} & 0.500 & 0.500 & 0.500 & 0.508 & 0.522  \\
116 & FOREX\_cadjpy-hour-High & \textbf{0.582} & 0.500 & 0.500 & 0.500 & 0.504 & 0.525  \\
155 & jungle\_chess\_endgame & 0.777 & 0.755 & 0.748 & 0.789 & 0.779 & \textbf{0.815}  \\
30 & bank & \textbf{0.794} & 0.760 & 0.759 & 0.781 & 0.778 & 0.721  \\
93 & electricity & 0.774 & 0.761 & 0.759 & 0.760 & 0.769 & \textbf{0.847}  \\
188 & mobile\_c36\_oversampling & 0.904 & 0.905 & 0.909 & 0.910 & 0.915 & \textbf{0.973}  \\
42 & BNG(cmc) & 0.537 & 0.526 & 0.513 & \textbf{0.546} & 0.527 & 0.521  \\
248 & shuttle & \textbf{0.997} & 0.981 & 0.984 & 0.914 & 0.939 & 0.880  \\

\multicolumn{2}{l}{\textit{Avg. \% Improv. over Baseline}} & -- & \colorbox{cyan!20}{\textbf{5.34\%}} & \colorbox{cyan!20}{\textbf{7.47\%}} & \colorbox{cyan!20}{\textbf{2.92\%}} & \colorbox{cyan!20}{\textbf{3.70\%}} & \colorbox{cyan!20}{\textbf{1.98\%}}  \\
\hline

\multicolumn{8}{c}{\textit{Large-d (14)}} \\ \hline
15 [VLS] & nomao & 0.932 & 0.927 & 0.929 & 0.950 & 0.930 & \textbf{0.951}  \\
121 & gas-drift & 0.974 & 0.968 & 0.978 & \textbf{0.983} & 0.982 & 0.981  \\
20 [VLS] & volkert & 0.399 & 0.347 & 0.358 & 0.415 & 0.423 & \textbf{0.509}  \\
177 & mfeat-factors & 0.948 & 0.949 & \textbf{0.958} & 0.955 & 0.948 & 0.803  \\
144 & Indian\_pines & \textbf{0.917} & 0.893 & 0.862 & 0.797 & 0.796 & 0.750  \\
181 & mfeat-pixel & 0.925 & 0.927 & \textbf{0.948} & 0.923 & 0.940 & 0.838  \\
171 & madeline & 0.622 & 0.568 & 0.556 & 0.620 & 0.631 & \textbf{0.708}  \\
8 [VLS] & fabert & 0.249 & 0.269 & 0.244 & \textbf{0.300} & 0.248 & 0.284  \\
10 [VLS] & gina\_agnostic & \textbf{0.859} & 0.796 & 0.836 & 0.814 & 0.839 & 0.809  \\
7 [VLS] & dilbert & 0.877 & 0.820 & 0.804 & \textbf{0.927} & 0.898 & 0.885  \\
12 [HD] & PCMAC\_1 & \textbf{0.835} & 0.659 & 0.564 & 0.507 & 0.526 & 0.484  \\
14 [HD] & RELATHE\_1 & \textbf{0.797} & 0.582 & 0.650 & 0.471 & 0.480 & 0.500  \\
3 [HD] & BASEHOCK\_1 & \textbf{0.935} & 0.639 & 0.624 & 0.675 & 0.760 & 0.665  \\
6 & gisette\_1 & \textbf{0.950} & 0.841 & 0.894 & 0.873 & 0.912 & 0.888  \\

\multicolumn{2}{l}{\textit{Avg. \% Improv. over Baseline}} & -- & \colorbox{cyan!20}{\textbf{11.31\%}} & \colorbox{cyan!20}{\textbf{11.91\%}} & \colorbox{cyan!20}{\textbf{12.33\%}} & \colorbox{cyan!20}{\textbf{11.31\%}} & \colorbox{cyan!20}{\textbf{13.34\%}}  \\ 


\midrule
\multicolumn{2}{l}{\textbf{\textit{Overall Avg. \% Improv. over Baseline}}} & -- & \colorbox{blue!20}{\textbf{8.13\%}} & \colorbox{blue!20}{\textbf{9.57\%}} & \colorbox{blue!20}{\textbf{7.32\%}} & \colorbox{blue!20}{\textbf{7.27\%}} & \colorbox{blue!20}{\textbf{7.28\%}}  \\
\multicolumn{2}{l}{\textbf{\textit{Overall Avg. \% $\Delta$ from Original (Table \ref{tab:og_results_tab})}}} & \colorbox{teal!40}{\textbf{3.43\%}} & \colorbox{teal!40}{\textbf{5.88\%}} & \colorbox{teal!40}{\textbf{4.81\%}} & \colorbox{teal!40}{\textbf{2.60\%}} & \colorbox{teal!40}{\textbf{0.65\%}} & \colorbox{teal!40}{\textbf{0.40\%}} \\
\bottomrule
\end{tabular}
}
\end{table}

\clearpage

\begin{table}[!t]
\centering
\caption{\textbf{TabPFNv1} Performance results using each method on individual datasets under the \textbf{HardCOp4DA} setting with \textbf{feature augmentation} $\text{F}_{\text{aug}}$. Best performance per dataset is shown in \textbf{bold}, with average percentage (\%) improvement by \method against each Baseline is highlighted \colorbox{blue!20}{in color}. Effect of feature augmentation is shown as the average percentage (\%) difference from performance on the original datasets, highlighted \colorbox{teal!40}{in green}.}
\label{tab:aug_feature_results_tab}
\resizebox{\textwidth}{!}{
\begin{tabular}{llcccccc}
\toprule
 & \textbf{Name} & \textbf{VIP\_COp} & \textbf{H1} & \textbf{H2} & \textbf{H3} & \textbf{O1} & \textbf{O2} \\

\textbf{ID} & \textbf{(Source)} & \textbf{(ours)} & \textbf{(Mean15)} & \textbf{(Ensem20)} & \textbf{(XXL)} & \textbf{(Kmeans)} & \textbf{(DT+TFM)} \\ \midrule

\multicolumn{8}{c}{\textit{small-d (16)}} \\ \hline
19 & Amazon\_employee\_access & \textbf{0.500} & \textbf{0.500} & \textbf{0.500} & \textbf{0.500} & \textbf{0.500} & \textbf{0.500}  \\
145 & INNHotelsGroup & \textbf{0.788} & 0.773 & 0.769 & 0.782 & 0.756 & 0.787  \\
41 & BNG(breast-w) & 0.979 & 0.979 & 0.978 & \textbf{0.983} & 0.948 & 0.916  \\
47 & BNG(tic-tac-toe) & \textbf{0.717} & 0.682 & 0.699 & 0.709 & 0.685 & 0.675  \\
55 & Click\_prediction & \textbf{0.569} & 0.492 & 0.500 & 0.500 & 0.488 & 0.471  \\
109 & FOREX\_audcad-hour-High & 0.556 & 0.512 & 0.500 & 0.657 & \textbf{0.676} & 0.599  \\
112 & FOREX\_audjpy-hour-High & 0.558 & 0.501 & 0.500 & 0.573 & \textbf{0.692} & 0.550  \\
113 & FOREX\_audsgd-hour-High & 0.530 & 0.500 & 0.500 & 0.500 & \textbf{0.669} & 0.534  \\
114 & FOREX\_audusd-hour-High & 0.521 & 0.500 & 0.500 & 0.500 & \textbf{0.671} & 0.531  \\
116 & FOREX\_cadjpy-hour-High & 0.552 & 0.502 & 0.500 & 0.506 & \textbf{0.674} & 0.537  \\
155 & jungle\_chess\_endgame & \textbf{0.757} & 0.650 & 0.609 & 0.667 & 0.688 & 0.697  \\
30 & bank & \textbf{0.764} & 0.603 & 0.561 & 0.609 & 0.672 & 0.602  \\
93 & electricity & 0.761 & 0.762 & 0.762 & \textbf{0.767} & 0.740 & 0.755  \\
188 & mobile\_c36\_oversampling & 0.909 & 0.906 & 0.904 & \textbf{0.920} & 0.890 & 0.904  \\
42 & BNG(cmc) & \textbf{0.535} & 0.513 & 0.525 & \textbf{0.535} & 0.523 & 0.503  \\
248 & shuttle & 0.575 & 0.499 & 0.445 & 0.474 & 0.688 & \textbf{0.743}  \\

\multicolumn{2}{l}{\textit{Avg. \% Improv. over Baseline}} & -- & \colorbox{cyan!20}{\textbf{7.86\%}} & \colorbox{cyan!20}{\textbf{9.65\%}} & \colorbox{cyan!20}{\textbf{4.68\%}} & \colorbox{cyan!20}{\textbf{-3.44\%}} & \colorbox{cyan!20}{\textbf{3.07\%}}  \\
\hline

\multicolumn{8}{c}{\textit{small-d (n$>$50K) (8)}} \\ \hline
147 & internet\_firewall & 0.497 & 0.416 & 0.497 & 0.250 & 0.249 & \textbf{0.509}  \\
52 & Cardiovascular-Disease & 0.705 & 0.700 & 0.696 & \textbf{0.708} & 0.654 & 0.687  \\
77 & Credit\_c & \textbf{0.661} & 0.582 & 0.613 & 0.601 & 0.600 & 0.649  \\
239 & SDSS17 & 0.952 & 0.939 & 0.940 & \textbf{0.960} & 0.932 & 0.916  \\
80 & dabetes\_us\_hospitals & 0.599 & 0.585 & 0.569 & \textbf{0.603} & 0.583 & 0.572  \\
15 & airline\_satisfaction & 0.887 & 0.894 & 0.880 & \textbf{0.930} & 0.892 & 0.893  \\
229 & Rain\_in\_Australia & \textbf{0.680} & 0.615 & 0.628 & 0.660 & 0.641 & 0.616  \\
8 & accelerometer & \textbf{0.687} & 0.438 & 0.243 & 0.447 & 0.400 & 0.471  \\

\multicolumn{2}{l}{\textit{Avg. \% Improv. over Baseline}} & -- & \colorbox{cyan!20}{\textbf{13.02\%}} & \colorbox{cyan!20}{\textbf{25.93\%}} & \colorbox{cyan!20}{\textbf{19.87\%}} & \colorbox{cyan!20}{\textbf{24.97\%}} & \colorbox{cyan!20}{\textbf{8.29\%}}  \\
\midrule
\multicolumn{2}{l}{\textbf{\textit{Overall Avg. \% Improv. over Baseline}}} & -- & \colorbox{blue!20}{\textbf{9.58\%}} & \colorbox{blue!20}{\textbf{15.05\%}} & \colorbox{blue!20}{\textbf{9.75\%}} & \colorbox{blue!20}{\textbf{6.02\%}} & \colorbox{blue!20}{\textbf{4.80\%}}  \\
\multicolumn{2}{l}{\textbf{\textit{Overall Avg. \% $\Delta$ from Original (Table \ref{tab:og_results_tab})}}} & \colorbox{teal!40}{\textbf{-1.43\%}} & \colorbox{teal!40}{\textbf{-1.86\%}} & \colorbox{teal!40}{\textbf{-1.80\%}} & \colorbox{teal!40}{\textbf{-3.88\%}} & \colorbox{teal!40}{\textbf{1.95\%}} & \colorbox{teal!40}{\textbf{-4.11\%}} \\
\bottomrule

\end{tabular}
}
\end{table}

\begin{figure}[h]
\vspace{1.15in}
\centering
\includegraphics[width=0.7\linewidth]{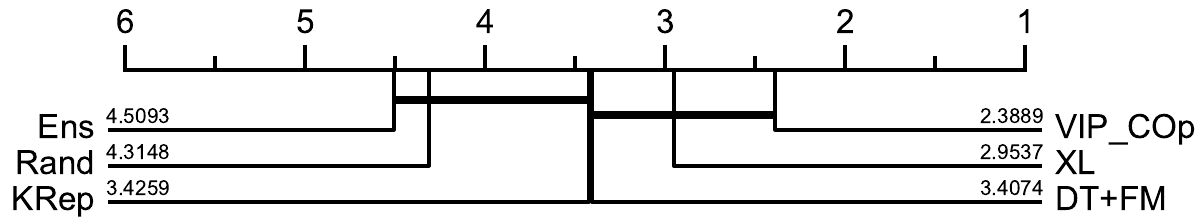}
\caption{CD diagram across datasets (numbers depict average rank) in the \textbf{HardCOp4DA} setting.}
\vspace{-0.05in}
\label{fig:augdemsar}
\end{figure}

\begin{table}[!th]
\centering
\caption{\textbf{TabPFNv1} Performance results using each method on individual datasets under the \textbf{HardCOp4DN} setting with \textbf{sample noising} $\text{S1}_{\text{noi}}$. Best performance per dataset is shown in \textbf{bold}, with average percentage (\%) improvement by \method against each Baseline is highlighted \colorbox{blue!20}{in color}. Effect of sample noise is shown as the average percentage (\%) difference from performance on the original datasets, highlighted \colorbox{teal!40}{in green}.}
\label{tab:noise_s1_results_tab}
\resizebox{\textwidth}{!}{
\begin{tabular}{llcccccc}
\toprule
 & \textbf{Name} & \textbf{VIP\_COp} & \textbf{H1} & \textbf{H2} & \textbf{H3} & \textbf{O1} & \textbf{O2} \\

\textbf{ID} & \textbf{(Source)} & \textbf{(ours)} & \textbf{(Mean15)} & \textbf{(Ensem20)} & \textbf{(XXL)} & \textbf{(Kmeans)} & \textbf{(DT+TFM)} \\ \midrule

\multicolumn{8}{c}{\textit{small-d (16)}} \\ \hline
19 & Amazon\_employee\_access & \textbf{0.500} & \textbf{0.500} & \textbf{0.500} & \textbf{0.500} & \textbf{0.500} & \textbf{0.500}  \\
145 & INNHotelsGroup & 0.786 & 0.677 & 0.693 & 0.730 & 0.734 & \textbf{0.841}  \\
41 & BNG(breast-w) & \textbf{0.981} & 0.961 & 0.966 & 0.967 & 0.968 & 0.966  \\
47 & BNG(tic-tac-toe) & 0.704 & 0.500 & 0.500 & 0.500 & 0.500 & \textbf{0.716}  \\
55 & Click\_prediction & \textbf{0.500} & \textbf{0.500} & \textbf{0.500} & \textbf{0.500} & \textbf{0.500} & \textbf{0.500}  \\
109 & FOREX\_audcad-hour-High & \textbf{0.673} & 0.502 & 0.500 & 0.500 & 0.497 & 0.563  \\
112 & FOREX\_audjpy-hour-High & \textbf{0.601} & 0.500 & 0.500 & 0.500 & 0.480 & 0.528  \\
113 & FOREX\_audsgd-hour-High & \textbf{0.542} & 0.500 & 0.500 & 0.500 & 0.500 & 0.517  \\
114 & FOREX\_audusd-hour-High & \textbf{0.531} & 0.500 & 0.500 & 0.500 & 0.500 & 0.512  \\
116 & FOREX\_cadjpy-hour-High & \textbf{0.572} & 0.500 & 0.500 & 0.500 & 0.500 & 0.520  \\
155 & jungle\_chess\_endgame & 0.723 & 0.463 & 0.516 & 0.459 & 0.510 & \textbf{0.763}  \\
30 & bank & \textbf{0.672} & 0.500 & 0.500 & 0.500 & 0.513 & 0.591  \\
93 & electricity & 0.767 & 0.709 & 0.711 & 0.732 & 0.723 & \textbf{0.794}  \\
188 & mobile\_c36\_oversampling & 0.911 & 0.874 & 0.880 & 0.899 & 0.884 & \textbf{0.930}  \\
42 & BNG(cmc) & 0.525 & 0.340 & 0.333 & 0.333 & 0.333 & \textbf{0.537}  \\
248 & shuttle & 0.634 & 0.432 & 0.285 & 0.428 & 0.296 & \textbf{0.748}  \\
\midrule
\multicolumn{2}{l}{\textbf{\textit{Overall Avg. \% Improv. over Baseline}}} & -- & \colorbox{blue!20}{\textbf{21.65\%}} & \colorbox{blue!20}{\textbf{25.35\%}} & \colorbox{blue!20}{\textbf{21.10\%}} & \colorbox{blue!20}{\textbf{24.54\%}} & \colorbox{blue!20}{\textbf{1.92\%}}  \\

\multicolumn{2}{l}{\textbf{\textit{Overall Avg. \% $\Delta$ from Original (Table \ref{tab:og_results_tab})}}} & \colorbox{teal!40}{\textbf{-1.44\%}} & \colorbox{teal!40}{\textbf{-11.21\%}} & \colorbox{teal!40}{\textbf{-13.03\%}} & \colorbox{teal!40}{\textbf{-13.78\%}} & \colorbox{teal!40}{\textbf{-15.54\%}} & \colorbox{teal!40}{\textbf{-0.80\%}} \\

\bottomrule

\end{tabular}
}
\end{table}
\begin{table}[!th]
\centering
\caption{\textbf{TabPFNv1} Performance results using each method on individual datasets under the \textbf{HardCOp4DN} setting with \textbf{sample noising} $\text{S2}_{\text{noi}}$. Best performance per dataset is shown in \textbf{bold}, with average percentage (\%) improvement by \method against each Baseline is highlighted \colorbox{blue!20}{in color}. Effect of sample noise is shown as the average percentage (\%) difference from performance on the original datasets, highlighted \colorbox{teal!40}{in green}.}
\label{tab:noise_s2_results_tab}
\resizebox{\textwidth}{!}{
\begin{tabular}{llcccccc}

\toprule
 & \textbf{Name} & \textbf{VIP\_COp} & \textbf{H1} & \textbf{H2} & \textbf{H3} & \textbf{O1} & \textbf{O2} \\

\textbf{ID} & \textbf{(Source)} & \textbf{(ours)} & \textbf{(Mean15)} & \textbf{(Ensem20)} & \textbf{(XXL)} & \textbf{(Kmeans)} & \textbf{(DT+TFM)} \\ \midrule

\multicolumn{8}{c}{\textit{small-d (16)}} \\ \hline
19 & Amazon\_employee\_access & \textbf{0.500} & \textbf{0.500} & \textbf{0.500} & \textbf{0.500} & \textbf{0.500} & \textbf{0.500}  \\
145 & INNHotelsGroup & 0.795 & 0.705 & 0.721 & 0.739 & 0.750 & \textbf{0.828}  \\
41 & BNG(breast-w) & \textbf{0.979} & 0.942 & 0.953 & 0.938 & 0.963 & 0.954  \\
47 & BNG(tic-tac-toe) & \textbf{0.710} & 0.500 & 0.500 & 0.500 & 0.577 & 0.703  \\
55 & Click\_prediction & \textbf{0.500} & \textbf{0.500} & \textbf{0.500} & \textbf{0.500} & \textbf{0.500} & \textbf{0.500}  \\
109 & FOREX\_audcad-hour-High & \textbf{0.682} & 0.502 & 0.500 & 0.500 & 0.500 & 0.544  \\
112 & FOREX\_audjpy-hour-High & \textbf{0.603} & 0.501 & 0.500 & 0.500 & 0.500 & 0.528  \\
113 & FOREX\_audsgd-hour-High & \textbf{0.561} & 0.500 & 0.500 & 0.500 & 0.528 & 0.520  \\
114 & FOREX\_audusd-hour-High & 0.505 & 0.500 & 0.500 & 0.500 & 0.500 & \textbf{0.513}  \\
116 & FOREX\_cadjpy-hour-High & \textbf{0.558} & 0.500 & 0.500 & 0.500 & 0.500 & 0.522  \\
155 & jungle\_chess\_endgame & 0.690 & 0.488 & 0.513 & 0.531 & 0.531 & \textbf{0.747}  \\
30 & bank & \textbf{0.686} & 0.508 & 0.500 & 0.513 & 0.545 & 0.567  \\
93 & electricity & 0.771 & 0.642 & 0.642 & 0.693 & 0.707 & \textbf{0.809}  \\
188 & mobile\_c36\_oversampling & 0.913 & 0.873 & 0.859 & 0.900 & 0.891 & \textbf{0.924}  \\
42 & BNG(cmc) & 0.530 & 0.346 & 0.333 & 0.389 & 0.357 & \textbf{0.533}  \\
248 & shuttle & \textbf{0.681} & 0.422 & 0.279 & 0.423 & 0.303 & 0.552  \\
\midrule
\multicolumn{2}{l}{\textbf{\textit{Overall Avg. \% Improv. over Baseline}}} & -- & \colorbox{blue!20}{\textbf{22.21\%}} & \colorbox{blue!20}{\textbf{27.38\%}} & \colorbox{blue!20}{\textbf{19.33\%}} & \colorbox{blue!20}{\textbf{21.69\%}} & \colorbox{blue!20}{\textbf{5.17\%}}  \\

\multicolumn{2}{l}{\textbf{\textit{Overall Avg. \% $\Delta$ from Original (Table \ref{tab:og_results_tab})}}} & \colorbox{teal!40}{\textbf{-1.09\%}} & \colorbox{teal!40}{\textbf{-11.39\%}} & \colorbox{teal!40}{\textbf{-13.71\%}} & \colorbox{teal!40}{\textbf{-12.80\%}} & \colorbox{teal!40}{\textbf{-13.46\%}} & \colorbox{teal!40}{\textbf{-3.51\%}} \\

\bottomrule

\end{tabular}
}
\end{table}

\clearpage
\begin{table}[!th]
\centering
\caption{\textbf{TabICLv2} Performance results using each method on individual datasets under the \textbf{HardCOp4DN} setting with $\text{S2}_{\text{noi}}+\text{S2}_{\text{noi}}$ \textbf{sample noising}. Best performance per dataset is shown in \textbf{bold}, with average percentage (\%) improvement by \method against each Baseline is highlighted \colorbox{blue!20}{in blue}. Effect of sample noise is shown as the average percentage (\%) difference from performance on the original datasets, highlighted \colorbox{teal!40}{in green}.}
\label{tab:sample_noise_s2s2_results_tab}
\resizebox{\textwidth}{!}{
\begin{tabular}{llcccccc}
\toprule
 & \textbf{Name} & \textbf{VIP\_COp} & \textbf{H1} & \textbf{H2} & \textbf{H3} & \textbf{O1} & \textbf{O2} \\

\textbf{ID} & \textbf{(Source)} & \textbf{(ours)} & \textbf{(Mean15)} & \textbf{(Ensem20)} & \textbf{(XXL)} & \textbf{(Kmeans)} & \textbf{(DT+TFM)} \\ \midrule

\multicolumn{8}{c}{\textit{small-d (16)}} \\ \hline

19 & Amazon\_employee\_access
& \textbf{0.506} & 0.500 & 0.500 & 0.500 & 0.500 & 0.500 \\

145 & INNHotelsGroup
& \textbf{0.810} & 0.808 & 0.808 & 0.808 & 0.808 & 0.808 \\

41 & BNG(breast-w)
& \textbf{0.975} & 0.967 & 0.967 & 0.967 & 0.967 & 0.967 \\

47 & BNG(tic-tac-toe)
& \textbf{0.691} & 0.684 & 0.684 & 0.684 & 0.684 & 0.684 \\

55 & Click\_prediction
& 0.500 & 0.500 & 0.500 & 0.500 & 0.500 & 0.500 \\

109 & FOREX\_audcad-hour-High
& \textbf{0.702} & 0.666 & 0.666 & 0.666 & 0.666 & 0.666 \\

112 & FOREX\_audjpy-hour-High
& \textbf{0.703} & 0.671 & 0.671 & 0.671 & 0.671 & 0.671 \\

113 & FOREX\_audsgd-hour-High
& \textbf{0.692} & 0.608 & 0.608 & 0.608 & 0.608 & 0.608 \\

114 & FOREX\_audusd-hour-High
& \textbf{0.691} & 0.548 & 0.548 & 0.548 & 0.548 & 0.548 \\

116 & FOREX\_cadjpy-hour-High
& \textbf{0.675} & 0.661 & 0.661 & 0.661 & 0.661 & 0.661 \\

155 & jungle\_chess\_endgame
& \textbf{0.522} & 0.487 & 0.487 & 0.487 & 0.487 & 0.487 \\

30 & bank
& \textbf{0.668} & 0.553 & 0.553 & 0.553 & 0.553 & 0.553 \\

93 & electricity
& \textbf{0.820} & 0.816 & 0.816 & 0.816 & 0.816 & 0.816 \\

188 & mobile\_c36\_oversampling
& \textbf{0.922} & 0.918 & 0.918 & 0.918 & 0.918 & 0.918 \\

42 & BNG(cmc)
& \textbf{0.498} & 0.477 & 0.477 & 0.477 & 0.477 & 0.477 \\

248 & shuttle
& \textbf{0.896} & 0.659 & 0.659 & 0.659 & 0.659 & 0.659 \\

\multicolumn{2}{l}{\textit{Avg. \% Improv. over Baseline}}
& -- 
& \colorbox{blue!20}{\textbf{7.82\%}}
& \colorbox{blue!20}{\textbf{7.82\%}}
& \colorbox{blue!20}{\textbf{7.82\%}}
& \colorbox{blue!20}{\textbf{7.82\%}}
& \colorbox{blue!20}{\textbf{7.82\%}} \\

\hline

\multicolumn{8}{c}{\textit{small-d ($n>50K$) (8)}} \\ \hline

147 & internet\_firewall
& \textbf{0.656} & 0.487 & 0.488 & 0.490 & 0.489 & 0.609 \\

52 & Cardiovascular-Disease
& \textbf{0.733} & 0.719 & 0.720 & 0.722 & 0.721 & 0.721 \\

77 & Credit\_c
& \textbf{0.759} & 0.708 & 0.720 & 0.710 & 0.707 & 0.710 \\

239 & SDSS17
& \textbf{0.971} & 0.734 & 0.741 & 0.737 & 0.732 & 0.739 \\

80 & dabetes\_us\_hospitals
& \textbf{0.678} & 0.621 & 0.623 & 0.619 & 0.619 & 0.622 \\

15 & airline\_satisfaction
& \textbf{0.957} & 0.808 & 0.819 & 0.809 & 0.835 & 0.811 \\

229 & Rain\_in\_Australia
& \textbf{0.685} & 0.642 & 0.654 & 0.649 & 0.648 & 0.658 \\

8 & accelerometer
& \textbf{0.794} & 0.721 & 0.747 & 0.743 & 0.730 & 0.788 \\

\multicolumn{2}{l}{\textit{Avg. \% Improv. over Baseline}}
& --
& \colorbox{blue!20}{\textbf{15.02\%}}
& \colorbox{blue!20}{\textbf{13.65\%}}
& \colorbox{blue!20}{\textbf{14.29\%}}
& \colorbox{blue!20}{\textbf{14.27\%}}
& \colorbox{blue!20}{\textbf{9.93\%}} \\

\midrule

\multicolumn{2}{l}{\textbf{\textit{Overall Avg. \% Improv. over Baseline}}}
& --
& \colorbox{blue!20}{\textbf{10.22\%}}
& \colorbox{blue!20}{\textbf{9.76\%}}
& \colorbox{blue!20}{\textbf{9.97\%}}
& \colorbox{blue!20}{\textbf{9.97\%}}
& \colorbox{blue!20}{\textbf{8.52\%}} \\

\multicolumn{2}{l}{\textbf{\textit{Overall Avg. \% $\Delta$ from Original (Table \ref{tab:og_results_tab})}}}
& \colorbox{teal!40}{\textbf{-6.64\%}}
& \colorbox{teal!40}{\textbf{-12.96\%}}
& \colorbox{teal!40}{\textbf{-12.59\%}}
& \colorbox{teal!40}{\textbf{-12.79\%}}
& \colorbox{teal!40}{\textbf{-12.80\%}}
& \colorbox{teal!40}{\textbf{-11.64\%}} \\

\bottomrule

\end{tabular}
}
\end{table}

\begin{table}[!th]
\centering
\caption{\textbf{TabPFNv1} Performance results using each method on individual datasets under the \textbf{HardCOp4DN} setting with \textbf{feature noising} $\text{F}_{\text{noi}}$ that mixes $\text{F1}_{\text{noi}}$+$\text{F2}_{\text{noi}}$. Best performance per dataset is shown in \textbf{bold}, with average percentage (\%) improvement by \method against each Baseline is highlighted \colorbox{blue!20}{in blue}. Effect of feature noise is shown as the average percentage (\%) difference from performance on the original datasets, highlighted \colorbox{teal!40}{in green}.}
\label{tab:noise_f1f2_results_tab}
\resizebox{\textwidth}{!}{
\begin{tabular}{llcccccc}
\toprule
 & \textbf{Name} & \textbf{VIP\_COp} & \textbf{H1} & \textbf{H2} & \textbf{H3} & \textbf{O1} & \textbf{O2} \\

\textbf{ID} & \textbf{(Source)} & \textbf{(ours)} & \textbf{(Mean15)} & \textbf{(Ensem20)} & \textbf{(XXL)} & \textbf{(Kmeans)} & \textbf{(DT+TFM)} \\ \midrule

\multicolumn{8}{c}{\textit{Large-d (14)}} \\ \hline
15 [VLS] & nomao & \textbf{0.924} & 0.583 & 0.647 & 0.593 & 0.660 & 0.690  \\
121 & gas-drift & \textbf{0.980} & 0.124 & 0.191 & 0.121 & 0.190 & 0.245  \\
20 [VLS] & volkert & \textbf{0.336} & 0.113 & 0.135 & 0.100 & 0.099 & 0.087  \\
177 & mfeat-factors & \textbf{0.963} & 0.111 & 0.223 & 0.098 & 0.240 & 0.118  \\
144 & Indian\_pines & \textbf{0.703} & 0.163 & 0.131 & 0.167 & 0.186 & 0.138  \\
181 & mfeat-pixel & \textbf{0.945} & 0.144 & 0.088 & 0.143 & 0.173 & 0.133  \\
171 & madeline & \textbf{0.614} & 0.529 & 0.524 & 0.533 & 0.513 & 0.542  \\
8 [VLS] & fabert & \textbf{0.247} & 0.154 & 0.148 & 0.156 & 0.149 & 0.156  \\
10 [VLS] & gina\_agnostic & \textbf{0.850} & 0.474 & 0.550 & 0.482 & 0.598 & 0.503  \\
7 [VLS] & dilbert & \textbf{0.857} & 0.263 & 0.276 & 0.249 & 0.328 & 0.241  \\
12 [HD] & PCMAC\_1 & \textbf{0.705} & 0.521 & 0.513 & 0.527 & 0.524 & 0.513  \\
14 [HD] & RELATHE\_1 & \textbf{0.668} & 0.487 & 0.486 & 0.487 & 0.516 & 0.487  \\
3 [HD] & BASEHOCK\_1 & \textbf{0.702} & 0.443 & 0.493 & 0.439 & 0.514 & 0.424  \\
6 & gisette\_1 & \textbf{0.901} & 0.463 & 0.457 & 0.500 & 0.611 & 0.500  \\
\midrule
\multicolumn{2}{l}{\textbf{\textit{Overall Avg. \% Improv. over Baseline}}} & -- & \colorbox{blue!20}{\textbf{229.18\%}} & \colorbox{blue!20}{\textbf{207.91\%}} & \colorbox{blue!20}{\textbf{240.81\%}} & \colorbox{blue!20}{\textbf{154.11\%}} & \colorbox{blue!20}{\textbf{212.34\%}}  \\

\multicolumn{2}{l}{\textbf{\textit{Overall Avg. \% $\Delta$ from Original (Table \ref{tab:og_results_tab})}}} & \colorbox{teal!40}{\textbf{-6.11\%}} & \colorbox{teal!40}{\textbf{-50.23\%}} & \colorbox{teal!40}{\textbf{-46.44\%}} & \colorbox{teal!40}{\textbf{-52.11\%}} & \colorbox{teal!40}{\textbf{-48.13\%}} & \colorbox{teal!40}{\textbf{-54.09\%}} \\

\bottomrule

\end{tabular}
}
\end{table}

\begin{figure}[h]
\vspace{1.15in}
\centering
\includegraphics[width=0.7\linewidth]{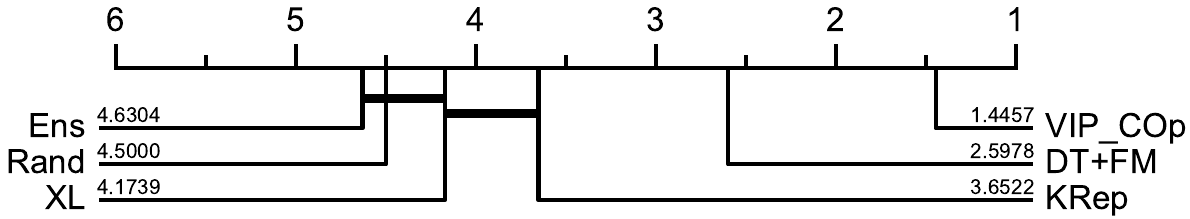}
\caption{CD diagram across datasets (numbers depict average rank) in the \textbf{HardCOp4DN} setting.}
\vspace{-0.05in}
\label{fig:noidemsar}
\vspace{-0.1in}
\end{figure}

\begin{figure*}[!t]
\centering
\begin{minipage}[t]{0.52\textwidth}
\centering

\begin{table}[H]
\centering
\caption{Pairwise comparison between each baseline and \method (TabPFN-v1 as backbone) based on $p$-value of permutation test over datasets across all HardCOp settings; Original, Data Augmentation (DA), Data Noise (DN). \method significantly outperforms the baselines in the majority of cases and overall, while none of the baselines consistently achieve competitive performance.}
\label{tab:pvals_v1}
\vspace{0.05in}
\scalebox{0.825}{
\begin{tabular}{lccc|cc}
\toprule
            & Rand & Ens & XL & KRep & DT+FM \\ 
            & \textbf{(H1)}  & \textbf{(H2)}  & \textbf{(H3)}  & \textbf{(O1)}  & \textbf{(O2)}  \\ 
\midrule
Original     & \textless1e-6 & \textless1e-6 & 0.002 & 0.012 & \colorbox{blue!15}{0.207} \\ 
\midrule
DA w/ $\text{S}_{\text{aug}}$  & \textless1e-6 & \textless1e-6 & 0.007 & 0.001 & 0.021 \\
DA w/ $\text{F}_{\text{aug}}$  & 1e-5 & \textless1e-6 & 0.014 & \colorbox{blue!15}{0.277} & 0.044 \\
\midrule
DN w/ $\text{S1}_{\text{noi}}$ & 2e-05 & 3e-05 & 6e-05 & 9e-05 & \colorbox{blue!15}{0.335} \\
DN w/ $\text{S2}_{\text{noi}}$ & 5e-05 & 3e-05 & 2e-05 & 5e-05 & \colorbox{blue!15}{0.052} \\
DN w/ $\text{F}_{\text{noi}}$  & 6e-05 & 9e-05 & 1e-05 & 2e-04 & 4e-05 \\
\hline 
\textbf{Overall} & \textbf{\textless 1e-6} & \textbf{\textless 1e-6} & \textbf{\textless 1e-6} & \textbf{\textless 1e-6} & \textbf{\textless 1e-6} \\
\bottomrule
\end{tabular}
}
\end{table}

\end{minipage}
\hfill
\begin{minipage}[t]{0.425\textwidth}
\centering

\begin{figure}[H]
\centering
\includegraphics[width=0.95\linewidth]{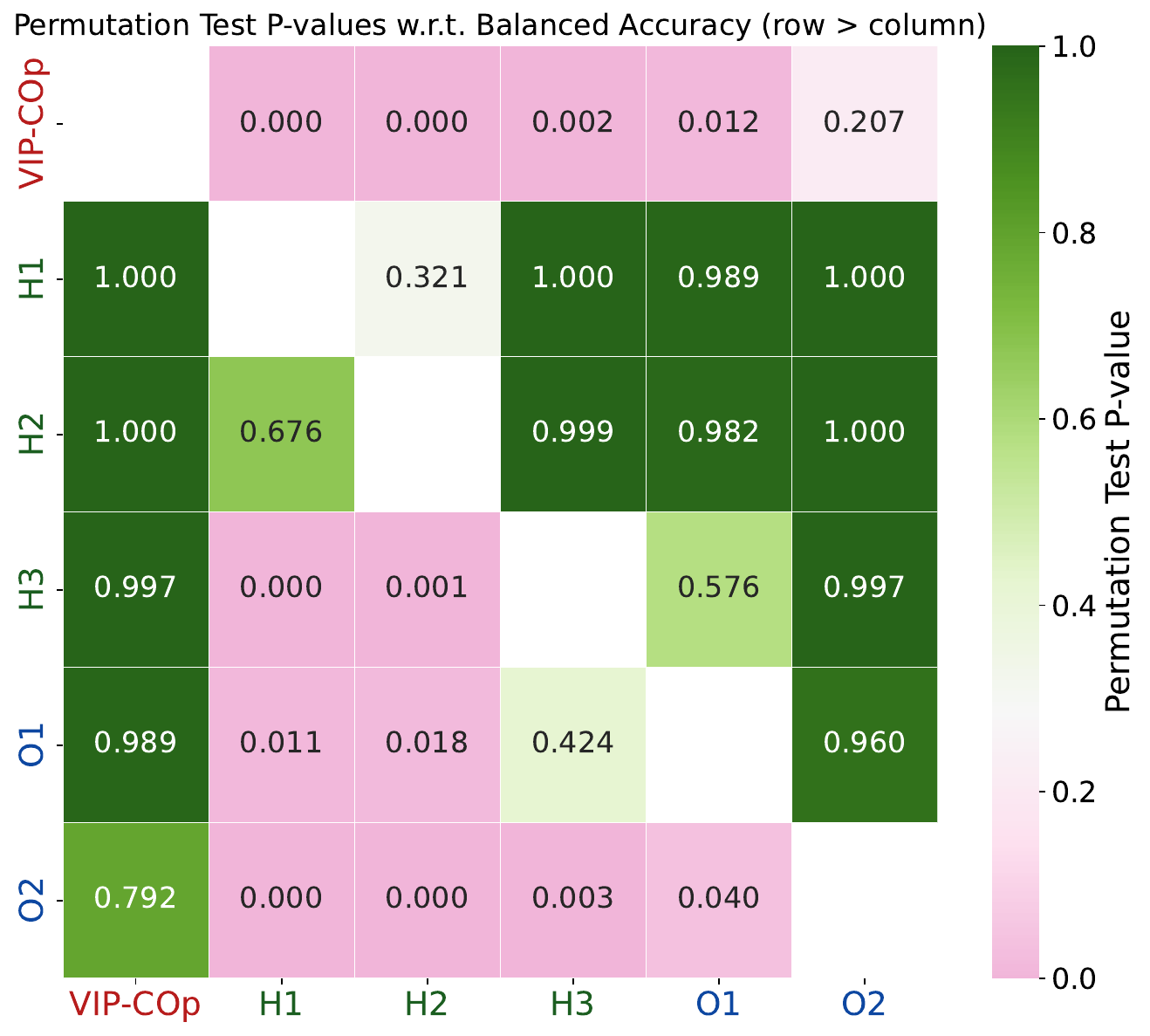}
\vspace{-0.05in}
\caption{Paired comparisons between all context optimization methods based on $p$-values of permutation tests over all (38) datasets in the {original HardCOp} setting, with TabPFNv1 as backbone.}
\label{fig:allpairs_v1}
\end{figure}

\end{minipage}
\vspace{0.1in}
\end{figure*}

\setlength{\tabcolsep}{3pt}
\renewcommand{\arraystretch}{0.92}
\small

\begin{longtable}{p{0.30\textwidth}cccccc}
\caption{\textbf{Performance results} using each method on individual datasets.}
\label{tab:odd_results_tab} \\

\toprule
\textbf{Name (Source)} & \textbf{VIP\_COp} & \textbf{H1} & \textbf{H2} & \textbf{H3} & \textbf{H4} & \textbf{O1} \\
 & \textbf{(Ours)} & \textbf{(Mean15)} & \textbf{(Ensem20)} & \textbf{(XXL)} & \textbf{(CMRS)} & \textbf{(KMeans)} \\
\midrule
\endhead

\midrule
\multicolumn{7}{r}{\textit{Continued on next page}} \\
\endfoot

\bottomrule
\endlastfoot

AbandonedMineLand & \textbf{0.749} & 0.479 & 0.477 & 0.472 & 0.492 & 0.534 \\
AdPerformance & \textbf{0.780} & 0.672 & 0.681 & 0.676 & 0.664 & 0.767 \\
AdRequestFraud & \textbf{0.756} & 0.528 & 0.524 & 0.502 & 0.629 & 0.589 \\
AgilePriority & \textbf{0.616} & 0.545 & 0.547 & 0.544 & 0.560 & 0.565 \\
AgriculturalYield & 0.998 & 0.993 & 0.999 & 0.998 & \textbf{1.000} & 0.995 \\
AircraftSensorAnomalies & 1.000 & 0.999 & 0.999 & 0.998 & 1.000 & 1.000 \\
AmbulanceDeployment & \textbf{0.868} & 0.848 & 0.846 & 0.859 & 0.703 & 0.220 \\
AnomalyClassification & 1.000 & 1.000 & 1.000 & 1.000 & 1.000 & 1.000 \\
ArtCollections & \textbf{0.820} & 0.584 & 0.615 & 0.604 & 0.631 & 0.664 \\
ArtSalesPlatform & \textbf{0.725} & 0.505 & 0.538 & 0.534 & 0.586 & 0.573 \\
ArxivProcessing & 0.875 & 0.690 & \textbf{0.973} & 0.882 & 0.964 & 0.741 \\
AstronomicalObservations & 1.000 & 1.000 & 1.000 & 1.000 & 1.000 & 0.997 \\
AutoencoderAnomaly & \textbf{0.549} & 0.515 & 0.514 & 0.526 & 0.517 & 0.526 \\
BackupSession & \textbf{0.908} & 0.858 & 0.854 & 0.900 & 0.871 & 0.849 \\
BankTransactions & \textbf{0.781} & 0.546 & 0.540 & 0.596 & 0.644 & 0.553 \\
BankruptcyStatus & \textbf{0.605} & 0.552 & 0.552 & 0.553 & 0.562 & 0.552 \\
BarExamPrediction & \textbf{0.794} & 0.602 & 0.620 & 0.630 & 0.700 & 0.684 \\
BaseballEvents & \textbf{0.702} & 0.627 & 0.631 & 0.617 & 0.620 & 0.616 \\
BearingFault & \textbf{0.989} & 0.948 & 0.959 & 0.943 & 0.975 & 0.892 \\
BettingIntegrity & 0.650 & \textbf{0.722} & 0.682 & 0.777 & 0.630 & 0.729 \\
BikeSharingTemperature & 0.988 & 0.963 & 0.979 & 0.967 & 0.995 & \textbf{0.998} \\
BioSequenceAnomaly & \textbf{0.853} & 0.602 & 0.587 & 0.580 & 0.692 & 0.673 \\
BiologicalClassifications & \textbf{0.974} & 0.868 & 0.905 & 0.864 & 0.969 & 0.904 \\
BirdMoult & \textbf{0.933} & 0.894 & 0.902 & 0.896 & 0.913 & 0.891 \\
BirdTracking & 0.999 & 0.999 & 0.999 & 0.999 & 0.999 & 0.999 \\
BlacklistSpam & \textbf{0.918} & 0.444 & 0.394 & 0.464 & 0.689 & 0.776 \\
BlockchainTransaction & \textbf{0.646} & 0.579 & 0.574 & 0.584 & 0.598 & 0.623 \\
BoliviaUnpaidLabor & \textbf{0.789} & 0.710 & 0.716 & 0.701 & 0.718 & 0.729 \\
BondMarketAnomalies & 0.927 & 0.941 & 0.943 & 0.909 & \textbf{0.974} & 0.907 \\
BuildingCalls & 1.000 & 1.000 & 1.000 & 1.000 & 1.000 & 1.000 \\
BuildingDamage & \textbf{0.621} & 0.547 & 0.550 & 0.539 & 0.543 & 0.514 \\
BuildingInspection & 0.912 & 0.874 & 0.881 & 0.859 & 0.887 & \textbf{0.917} \\
BuildingInspections & 0.743 & 0.506 & 0.627 & 0.466 & 0.771 & \textbf{0.772} \\
BuildingPoints & 1.000 & 1.000 & 1.000 & 1.000 & 1.000 & 1.000 \\
BusinessFraud & \textbf{0.543} & 0.487 & 0.485 & 0.483 & 0.506 & 0.483 \\
BusinessInspections & 0.443 & 0.479 & 0.471 & 0.455 & 0.478 & \textbf{0.498} \\
BusinessProcessActivity & \textbf{0.683} & 0.611 & 0.567 & 0.529 & 0.610 & 0.556 \\
CampaignOutcome & 1.000 & 1.000 & 0.983 & 1.000 & 1.000 & 1.000 \\
CancerTreatment & 0.568 & 0.555 & 0.559 & 0.553 & \textbf{0.574} & 0.574 \\
CandidateVetting & \textbf{0.568} & 0.534 & 0.535 & 0.537 & 0.534 & 0.534 \\
CarMakeAnomaly & \textbf{0.781} & 0.662 & 0.695 & 0.717 & 0.675 & 0.741 \\
CardFraudDetection & 0.965 & 0.949 & 0.944 & 0.933 & \textbf{0.967} & 0.951 \\
CardioCatchBMI & 0.998 & 0.995 & 0.983 & 0.992 & 0.998 & 0.998 \\
CarrefourSalesDefects & \textbf{0.623} & 0.563 & 0.579 & 0.562 & 0.579 & 0.578 \\
CasePriority & 0.997 & 0.990 & 0.997 & 0.996 & 0.996 & 0.993 \\
ChecksumValidation & 0.950 & 0.949 & 0.933 & 0.948 & 0.950 & 0.949 \\
ChicagoFoodInspection & \textbf{0.606} & 0.582 & 0.587 & 0.574 & 0.585 & 0.577 \\
ChronicDiseaseStats & 0.999 & 0.999 & 0.997 & 0.999 & \textbf{1.000} & 0.999 \\
CodeDefects & \textbf{0.666} & 0.623 & 0.595 & 0.613 & 0.653 & 0.614 \\
CogLoadReaction & 0.997 & 0.997 & 0.999 & 0.995 & \textbf{0.999} & 0.983 \\
ContactTypeMonitoring & 0.874 & 0.957 & 0.949 & \textbf{0.998} & 0.873 & 0.997 \\
ConversationAnomalies & 0.871 & 0.850 & 0.887 & 0.913 & \textbf{0.920} & 0.913 \\
CoralDisease & \textbf{0.871} & 0.515 & 0.507 & 0.474 & 0.614 & 0.427 \\
CountyHealthMetrics & 0.926 & 0.923 & 0.916 & \textbf{0.943} & 0.904 & 0.909 \\
CourseEvaluation & 0.619 & 0.172 & 0.326 & 0.183 & 0.214 & \textbf{0.757} \\
CourseRegistration & \textbf{0.693} & 0.555 & 0.572 & 0.520 & 0.678 & 0.569 \\
CovidEpidemiology & 0.655 & 0.644 & 0.699 & 0.641 & 0.663 & \textbf{0.860} \\
CovidTestingData & \textbf{0.931} & 0.809 & 0.807 & 0.841 & 0.880 & 0.720 \\
CovtypeAnomaly & \textbf{0.998} & 0.988 & 0.995 & 0.987 & 0.997 & 0.988 \\
CreditCardFraud & 0.789 & 0.770 & 0.776 & 0.771 & 0.767 & \textbf{0.798} \\
CreditDefault & \textbf{0.638} & 0.585 & 0.585 & 0.578 & 0.586 & 0.554 \\
CreditRiskModel & 0.677 & 0.462 & 0.413 & 0.549 & 0.478 & \textbf{0.768} \\
CrimeAnomalies & \textbf{0.530} & 0.522 & 0.521 & 0.523 & 0.528 & 0.510 \\
CrimeClearanceStatus & 0.582 & 0.551 & 0.568 & 0.565 & 0.581 & \textbf{0.670} \\
CrimeFraudDetection & 0.998 & 0.998 & 0.999 & 0.997 & \textbf{0.999} & 0.962 \\
CropProduction & \textbf{0.948} & 0.801 & 0.816 & 0.840 & 0.879 & 0.768 \\
CustomerCampaignOutcome & 1.000 & 0.985 & 0.972 & 0.966 & 1.000 & 1.000 \\
CustomerOrderPriority & \textbf{0.945} & 0.918 & 0.930 & 0.920 & 0.926 & 0.904 \\
CustomerServiceTransactions & \textbf{0.591} & 0.369 & 0.422 & 0.339 & 0.433 & 0.567 \\
CybersecurityVulnerability & 1.000 & 0.974 & 0.972 & 0.980 & 0.997 & 1.000 \\
DallasBankGeography & 0.569 & \textbf{0.581} & 0.535 & 0.566 & 0.579 & 0.456 \\
DamageFunctionSources & 0.500 & 0.500 & 0.492 & 0.500 & 0.500 & 0.500 \\
DataValidationEntries & \textbf{0.623} & 0.584 & 0.588 & 0.591 & 0.596 & 0.579 \\
DeathCauses & \textbf{0.648} & 0.574 & 0.576 & 0.571 & 0.593 & 0.571 \\
DeliveryRootCause & 0.595 & 0.566 & 0.558 & 0.575 & \textbf{0.595} & 0.517 \\
DentalConditionAnomaly & \textbf{0.547} & 0.523 & 0.517 & 0.522 & 0.526 & 0.513 \\
DiabetesSpecialty & 0.849 & 0.827 & 0.851 & \textbf{0.867} & 0.844 & 0.743 \\
DiamondClarity & \textbf{0.932} & 0.820 & 0.833 & 0.827 & 0.849 & 0.800 \\
DifferentialExpression & 0.500 & \textbf{1.000} & 0.860 & \textbf{1.000} & 0.500 & \textbf{1.000} \\
DisasterIncidentType & 0.327 & 0.370 & 0.358 & 0.171 & 0.394 & \textbf{0.509} \\
DoctorVisits & 0.998 & 0.969 & 0.978 & 0.955 & 0.999 & \textbf{1.000} \\
DragonFlySystem & 0.533 & 0.501 & 0.524 & 0.515 & 0.524 & \textbf{0.564} \\
DurhamVotersAge & 0.500 & 0.487 & 0.595 & 0.500 & 0.477 & \textbf{0.962} \\
EcommerceOrderPriority & \textbf{0.686} & 0.568 & 0.557 & 0.571 & 0.574 & 0.534 \\
ElectricBuildingUsage & \textbf{0.719} & 0.530 & 0.518 & 0.522 & 0.514 & 0.645 \\
ElectronicTesting & \textbf{0.869} & 0.841 & 0.852 & 0.846 & 0.855 & 0.835 \\
EmailPhishingDetection & \textbf{0.982} & 0.937 & 0.953 & 0.925 & 0.976 & 0.950 \\
EmotionalIncidents & \textbf{0.629} & 0.477 & 0.472 & 0.457 & 0.503 & 0.456 \\
EmployeeAttrition & 0.898 & 0.761 & 0.807 & 0.801 & 0.775 & \textbf{0.923} \\
EmployeeLocationAnomalies & \textbf{0.550} & 0.531 & 0.528 & 0.531 & 0.524 & 0.532 \\
EndangeredSpeciesClassification & \textbf{0.658} & 0.561 & 0.537 & 0.524 & 0.595 & 0.510 \\
GPUResourceUsage & \textbf{0.721} & 0.426 & 0.479 & 0.441 & 0.553 & 0.503 \\
GeographicSeverity & 0.946 & 0.772 & 0.899 & 0.932 & 0.964 & \textbf{0.954} \\
GeologicalLithology & 0.535 & 0.502 & 0.604 & 0.464 & 0.734 & \textbf{0.740} \\
GeospatialCrimeResolution & \textbf{0.732} & 0.578 & 0.585 & 0.625 & 0.627 & 0.675 \\
GlobalSharkAttacks & 0.616 & 0.508 & 0.560 & 0.500 & 0.616 & \textbf{0.657} \\
GlobalTerrorismAttack & \textbf{0.665} & 0.502 & 0.552 & 0.519 & 0.571 & 0.642 \\
HealthMetricsAnomalies & 0.505 & 0.501 & 0.500 & 0.502 & \textbf{0.507} & 0.498 \\
HealthPulse & \textbf{0.672} & 0.606 & 0.616 & 0.590 & 0.616 & 0.622 \\
HealthcareMessages & \textbf{0.774} & 0.677 & 0.726 & 0.560 & 0.748 & 0.688 \\
HockeyGameplay & 0.511 & 0.505 & 0.508 & 0.508 & 0.511 & \textbf{0.521} \\
HospitalSeverity & 0.511 & 0.507 & 0.507 & 0.503 & 0.505 & \textbf{0.521} \\
HousePriority & 0.500 & 0.500 & 0.498 & 0.500 & 0.500 & 0.499 \\
HousingConstruction & 0.775 & 0.767 & 0.783 & 0.758 & \textbf{0.805} & 0.749 \\
HuntingCriteria & 0.979 & 0.953 & 0.952 & 0.950 & \textbf{0.982} & 0.981 \\
ImageCategoryAnomaly & \textbf{0.607} & 0.508 & 0.489 & 0.512 & 0.514 & 0.456 \\
IndustrialEquipment & 0.998 & 0.988 & 0.995 & 0.998 & \textbf{0.999} & 0.998 \\
IndustrialFanVibration & \textbf{0.984} & 0.982 & 0.983 & 0.983 & 0.984 & 0.982 \\
ItemIncineration & 0.663 & 0.643 & 0.653 & 0.660 & 0.663 & \textbf{0.770} \\
JobStatus & \textbf{0.998} & 0.998 & 0.998 & 0.995 & 0.997 & 0.993 \\
JobStatusMonitoring & \textbf{0.879} & 0.796 & 0.786 & 0.781 & 0.823 & 0.796 \\
KidneyDisease & 0.998 & 0.992 & 0.991 & 0.994 & \textbf{1.000} & 0.999 \\
LabTestAbnormality & 0.683 & \textbf{0.732} & 0.683 & 0.628 & 0.671 & 0.693 \\
LandParcels & \textbf{0.739} & 0.504 & 0.582 & 0.408 & 0.724 & 0.697 \\
LandShapeTrade & 0.551 & 0.536 & 0.542 & 0.528 & 0.547 & \textbf{0.555} \\
LanguageAnomaly & \textbf{0.604} & 0.577 & 0.573 & 0.580 & 0.579 & 0.567 \\
LanguageLocationAnomaly & \textbf{0.904} & 0.385 & 0.553 & 0.494 & 0.487 & 0.870 \\
LargeWoodMeasurement & 1.000 & 0.999 & 1.000 & 0.999 & 1.000 & 0.999 \\
LaunchStatus & \textbf{0.734} & 0.580 & 0.608 & 0.566 & 0.670 & 0.640 \\
LegalChargeAnomalies & \textbf{0.979} & 0.923 & 0.940 & 0.901 & 0.969 & 0.978 \\
LegalEducation & \textbf{0.679} & 0.592 & 0.587 & 0.567 & 0.617 & 0.609 \\
LoanCreditScore & \textbf{0.710} & 0.631 & 0.636 & 0.628 & 0.657 & 0.630 \\
LoanFraudDetection & 0.756 & 0.736 & 0.760 & 0.692 & \textbf{0.769} & 0.692 \\
LoyaltyFraud & \textbf{0.878} & 0.830 & 0.846 & 0.823 & 0.845 & 0.790 \\
MaritimeVessel & 0.990 & 0.989 & \textbf{0.997} & 0.996 & 0.987 & 0.974 \\
MaterialFlow & \textbf{0.791} & 0.674 & 0.679 & 0.667 & 0.692 & 0.707 \\
MechanicalTesting & \textbf{0.942} & 0.680 & 0.735 & 0.670 & 0.755 & 0.865 \\
MedicalClassificationAnomalies & 0.584 & 0.756 & 0.863 & 0.771 & \textbf{0.991} & 0.987 \\
MicrostructureDefect & \textbf{0.744} & 0.717 & 0.724 & 0.707 & 0.743 & 0.679 \\
MineralComposition & 0.998 & 0.996 & 0.997 & 0.993 & 0.998 & 0.988 \\
MobileMoneyTransactions & \textbf{0.615} & 0.543 & 0.548 & 0.544 & 0.558 & 0.539 \\
MojaveDune & 0.995 & 0.995 & \textbf{0.997} & 0.994 & 0.995 & 0.991 \\
MouseInteractionPhase & 0.927 & 0.832 & 0.879 & 0.834 & 0.881 & \textbf{0.996} \\
MovieGenreAnomalies & \textbf{0.750} & 0.670 & 0.677 & 0.674 & 0.684 & 0.684 \\
MusicFamiliarity & \textbf{0.546} & 0.525 & 0.527 & 0.522 & 0.536 & 0.532 \\
NestSuccess & 0.999 & 0.997 & 0.999 & 0.997 & 0.998 & 0.997 \\
NetworkMalware & \textbf{0.796} & 0.690 & 0.695 & 0.687 & 0.725 & 0.693 \\
NetworkResponseAnomalies & 1.000 & 0.908 & 0.939 & 0.894 & 1.000 & 1.000 \\
NetworkThreat & 0.696 & 0.617 & 0.604 & 0.622 & \textbf{0.737} & 0.439 \\
NetworkTrafficDOS & 0.999 & 0.985 & 0.977 & 0.960 & 0.999 & 0.996 \\
NetworkTrafficSpam & 0.829 & 0.322 & 0.598 & 0.398 & \textbf{0.950} & 0.760 \\
NewbornReferral & \textbf{0.505} & 0.426 & 0.416 & 0.349 & 0.453 & 0.467 \\
NonComplianceInspection & 0.655 & 0.593 & 0.613 & 0.610 & 0.623 & \textbf{0.670} \\
OceanicStatus & 0.998 & 0.995 & 0.998 & 0.993 & 0.997 & 0.995 \\
OptimizationError & 0.998 & 0.999 & \textbf{1.000} & 0.999 & 0.999 & 0.997 \\
OrderTransmission & \textbf{0.493} & 0.467 & 0.470 & 0.471 & 0.479 & 0.465 \\
OzoneLevels & 0.987 & 0.899 & 0.918 & 0.887 & \textbf{0.988} & 0.962 \\
PatientReadmission & \textbf{0.581} & 0.526 & 0.528 & 0.527 & 0.541 & 0.515 \\
PhotoAnomalyDetection & \textbf{0.841} & 0.366 & 0.376 & 0.368 & 0.387 & 0.407 \\
PlayerMatchEvents & \textbf{0.647} & 0.621 & 0.624 & 0.620 & 0.624 & 0.635 \\
PokemonTypeAnomaly & 0.953 & 0.935 & 0.939 & 0.940 & 0.955 & \textbf{0.968} \\
PoliceFraudIncidents & 0.486 & 0.388 & 0.420 & 0.391 & 0.424 & \textbf{0.489} \\
PoliticalInterestMisinformation & \textbf{0.941} & 0.904 & 0.902 & 0.899 & 0.918 & 0.898 \\
PopulationComparison & \textbf{0.975} & 0.241 & 0.323 & 0.045 & 0.161 & 0.872 \\
PostcodeValidation & 0.481 & 0.555 & 0.644 & \textbf{0.670} & 0.523 & 0.481 \\
PrisonBooking & \textbf{0.522} & 0.485 & 0.487 & 0.491 & 0.501 & 0.509 \\
ProcessState & \textbf{0.626} & 0.533 & 0.532 & 0.536 & 0.541 & 0.542 \\
ProductionRunStatus & \textbf{0.987} & 0.809 & 0.872 & 0.792 & 0.975 & 0.955 \\
ProgrammingAnomalies & 0.528 & 0.523 & 0.526 & 0.522 & \textbf{0.548} & 0.532 \\
ProjectSeverity & 0.595 & 0.520 & 0.566 & 0.522 & 0.525 & \textbf{0.722} \\
PropertyCondition & \textbf{0.491} & 0.465 & 0.465 & 0.469 & 0.478 & 0.458 \\
ProteinCodonAnalysis & \textbf{0.928} & 0.582 & 0.501 & 0.616 & 0.688 & 0.619 \\
ProtestEvents & 0.618 & 0.501 & 0.558 & 0.500 & 0.516 & \textbf{0.732} \\
ProtestViolence & \textbf{0.809} & 0.677 & 0.696 & 0.614 & 0.773 & 0.740 \\
PsychTrialAccuracy & \textbf{0.559} & 0.505 & 0.516 & 0.502 & 0.522 & 0.551 \\
PsychologicalTrialValidity & 0.833 & 0.612 & 0.699 & 0.612 & 0.743 & \textbf{0.849} \\
PsychologyExperiment & \textbf{0.553} & 0.538 & 0.539 & 0.550 & 0.548 & 0.544 \\
RealEstateCondition & \textbf{0.717} & 0.637 & 0.652 & 0.636 & 0.658 & 0.669 \\
RealEstateProperty & \textbf{0.912} & 0.757 & 0.765 & 0.757 & 0.790 & 0.734 \\
RegionalPriceMonitoring & 0.985 & 0.984 & 0.976 & 0.984 & 0.985 & \textbf{1.000} \\
ReportResolution & \textbf{0.646} & 0.624 & 0.619 & 0.635 & 0.643 & 0.613 \\
ResidueLigandClassification & 0.871 & 0.866 & 0.876 & 0.818 & 0.865 & \textbf{0.889} \\
RestaurantViolation & \textbf{0.541} & 0.370 & 0.391 & 0.363 & 0.450 & 0.381 \\
RiskVulnerability & 0.500 & 0.500 & 0.494 & 0.500 & 0.500 & 0.500 \\
RoutingAnomaly & 0.983 & 0.982 & 0.982 & 0.983 & 0.983 & \textbf{0.992} \\
SalesOrderPriority & \textbf{0.609} & 0.551 & 0.545 & 0.549 & 0.576 & 0.532 \\
SalesTaskStatus & \textbf{0.748} & 0.559 & 0.567 & 0.555 & 0.621 & 0.583 \\
SatallaxBenchmark & \textbf{0.799} & 0.572 & 0.638 & 0.502 & 0.747 & 0.773 \\
SaudiWeatherAnomalies & 0.279 & 0.298 & 0.285 & \textbf{0.643} & 0.194 & 0.368 \\
SchedulerJobState & 0.993 & 0.967 & 0.961 & 0.963 & 0.992 & \textbf{0.995} \\
SensorimotorAdaption & \textbf{0.993} & 0.856 & 0.849 & 0.849 & 0.895 & 0.881 \\
SentencingFraud & \textbf{0.811} & 0.699 & 0.705 & 0.694 & 0.715 & 0.742 \\
ServerLoadTest & 0.500 & 0.500 & 0.534 & 0.500 & 0.500 & \textbf{0.650} \\
ServiceErrorLocation & \textbf{0.615} & 0.584 & 0.588 & 0.601 & 0.595 & 0.600 \\
ShapeTrackingAnomaly & 0.738 & 0.750 & 0.766 & 0.770 & 0.728 & \textbf{0.806} \\
ShellfishQuality & \textbf{0.567} & 0.504 & 0.486 & 0.512 & 0.499 & 0.457 \\
ShuttleOutlier & 0.990 & 0.987 & 0.992 & 0.987 & \textbf{1.000} & 0.979 \\
SierraLeoneWaterPoints & \textbf{0.682} & 0.554 & 0.552 & 0.575 & 0.596 & 0.581 \\
SmallBusinessLoans & \textbf{0.993} & 0.916 & 0.943 & 0.912 & 0.964 & 0.899 \\
SnowDepthMonitoring & \textbf{0.642} & 0.620 & 0.619 & 0.621 & 0.632 & 0.625 \\
SoccerActions & \textbf{0.697} & 0.689 & 0.694 & 0.685 & 0.688 & 0.664 \\
SoccerMatchAnomalies & 0.983 & 0.995 & 0.996 & \textbf{0.999} & 0.985 & 0.992 \\
SoilFloodProperties & 0.899 & 0.776 & 0.852 & 0.848 & 0.887 & \textbf{0.953} \\
SpamDetection & \textbf{0.597} & 0.440 & 0.447 & 0.435 & 0.508 & 0.418 \\
SpeciesObservation & 0.992 & 0.982 & 0.993 & 0.990 & 0.981 & 0.969 \\
SpecimenStatus & \textbf{0.994} & 0.772 & 0.832 & 0.731 & 0.983 & 0.836 \\
StatisticalTestResults & 1.000 & 0.998 & 1.000 & 0.999 & 1.000 & 0.998 \\
StoreCrowd & 0.500 & 0.500 & 0.493 & 0.500 & 0.500 & 0.500 \\
StormBehavior & \textbf{0.590} & 0.336 & 0.354 & 0.354 & 0.387 & 0.343 \\
StudentAssessment & 0.542 & 0.521 & 0.517 & 0.520 & 0.529 & \textbf{0.551} \\
StudentEnrollment & \textbf{0.732} & 0.547 & 0.553 & 0.538 & 0.592 & 0.565 \\
StudentProgramAnomaly & \textbf{0.687} & 0.427 & 0.420 & 0.388 & 0.525 & 0.357 \\
SummaryAnomalies & \textbf{0.518} & 0.501 & 0.494 & 0.502 & 0.507 & 0.503 \\
SuperStoreOrderPriority & \textbf{0.576} & 0.533 & 0.542 & 0.530 & 0.538 & 0.490 \\
SystemFailure & \textbf{0.713} & 0.575 & 0.578 & 0.573 & 0.602 & 0.546 \\
SystemFailureMonitoring & 0.740 & 0.709 & 0.762 & 0.705 & 0.736 & \textbf{0.910} \\
SystemLogs & \textbf{0.398} & 0.306 & 0.330 & 0.343 & 0.319 & 0.372 \\
SystemProcessAnomalies & \textbf{0.615} & 0.557 & 0.568 & 0.613 & 0.613 & 0.498 \\
SystemStatusAnomaly & \textbf{0.821} & 0.435 & 0.585 & 0.460 & 0.725 & 0.744 \\
TechnicalAnomalies & \textbf{0.855} & 0.824 & 0.817 & 0.822 & 0.836 & 0.801 \\
TelecomMarketing & 0.999 & 0.992 & 0.981 & 0.994 & 0.998 & \textbf{1.000} \\
TestCycleAnomalies & \textbf{0.903} & 0.729 & 0.719 & 0.625 & 0.797 & 0.739 \\
TVShowAnomalies & \textbf{0.732} & 0.472 & 0.475 & 0.466 & 0.510 & 0.488 \\
TicketPriority & \textbf{0.675} & 0.559 & 0.557 & 0.574 & 0.565 & 0.650 \\
TorontoPoliceContacts & 0.438 & 0.493 & 0.486 & 0.482 & \textbf{0.512} & 0.442 \\
TrafficIncident & \textbf{0.790} & 0.609 & 0.640 & 0.609 & 0.664 & 0.679 \\
TrafficStopAnomalies & \textbf{0.819} & 0.525 & 0.520 & 0.508 & 0.755 & 0.297 \\
TransactionFeeAnomalies & \textbf{0.649} & 0.634 & 0.637 & 0.634 & 0.641 & 0.643 \\
TransactionService & 0.992 & 0.982 & 0.982 & 0.991 & \textbf{0.998} & 0.993 \\
TrialConditionAnomaly & \textbf{0.572} & 0.536 & 0.536 & 0.523 & 0.554 & 0.541 \\
TurnaroundValidation & \textbf{0.667} & 0.562 & 0.559 & 0.565 & 0.569 & 0.558 \\
TwitchStreamAnomalies & \textbf{0.926} & 0.827 & 0.852 & 0.828 & 0.889 & 0.878 \\
UserEventAnomalies & 0.987 & 0.961 & 0.939 & 0.969 & \textbf{0.989} & 0.861 \\
VehicleNetworkTraffic & \textbf{0.927} & 0.799 & 0.790 & 0.770 & 0.903 & 0.871 \\
VenezuelaBeneficiaries & 0.500 & 0.497 & 0.614 & 0.581 & \textbf{0.745} & 0.639 \\
VideoGameViolence & 0.500 & 0.500 & 0.581 & 0.500 & 0.500 & \textbf{0.836} \\
VolcanicAnomalies & \textbf{0.908} & 0.516 & 0.566 & 0.512 & 0.716 & 0.563 \\
VolleyballMatch & \textbf{0.821} & 0.654 & 0.666 & 0.657 & 0.656 & 0.771 \\
VoteOutcome & \textbf{0.981} & 0.970 & 0.976 & 0.969 & 0.975 & 0.950 \\
WaterMeterAnomaly & \textbf{0.947} & 0.891 & 0.889 & 0.891 & 0.907 & 0.881 \\
WaterPointFunctionality & \textbf{0.674} & 0.560 & 0.563 & 0.571 & 0.556 & 0.586 \\
WaterQualityAssessment & 0.999 & 0.999 & \textbf{1.000} & 0.999 & 1.000 & 1.000 \\
WeatherEvents & \textbf{0.940} & 0.926 & 0.923 & 0.927 & 0.935 & 0.921 \\
WikipediaVandalism & 0.514 & 0.537 & 0.508 & 0.555 & 0.525 & 0.440 \\
WindTurbineFault & \textbf{0.598} & 0.511 & 0.521 & 0.510 & 0.532 & 0.535 \\
WindmillAnomaly & \textbf{0.957} & 0.772 & 0.783 & 0.774 & 0.804 & 0.778 \\
WindowsEditionAnomaly & 0.883 & 0.934 & 0.950 & 0.929 & \textbf{0.961} & 0.935 \\
WorkOvertimePatterns & 0.479 & 0.398 & 0.516 & 0.400 & 0.484 & \textbf{0.816} \\
YouTubeFaceDetection & \textbf{0.605} & 0.591 & 0.574 & 0.592 & 0.602 & 0.534 \\

\midrule
\textbf{Avg. \% Improv. over Baseline} &
- &
\colorbox{blue!20}{\textbf{16.37\%}} &
\colorbox{blue!20}{\textbf{12.41\%}} &
\colorbox{blue!20}{\textbf{23.75\%}} &
\colorbox{blue!20}{\textbf{10.01\%}} &
\colorbox{blue!20}{\textbf{9.13\%}} \\

\end{longtable}
\begin{figure*}[t]
\centering

\begin{subfigure}[t]{0.32\textwidth}
\centering
\includegraphics[width=\linewidth]{FIG/TIME_RELATHE_1.pdf}
\caption{RELATHE\_1}
\end{subfigure}
\hfill
\begin{subfigure}[t]{0.32\textwidth}
\centering
\includegraphics[width=\linewidth]{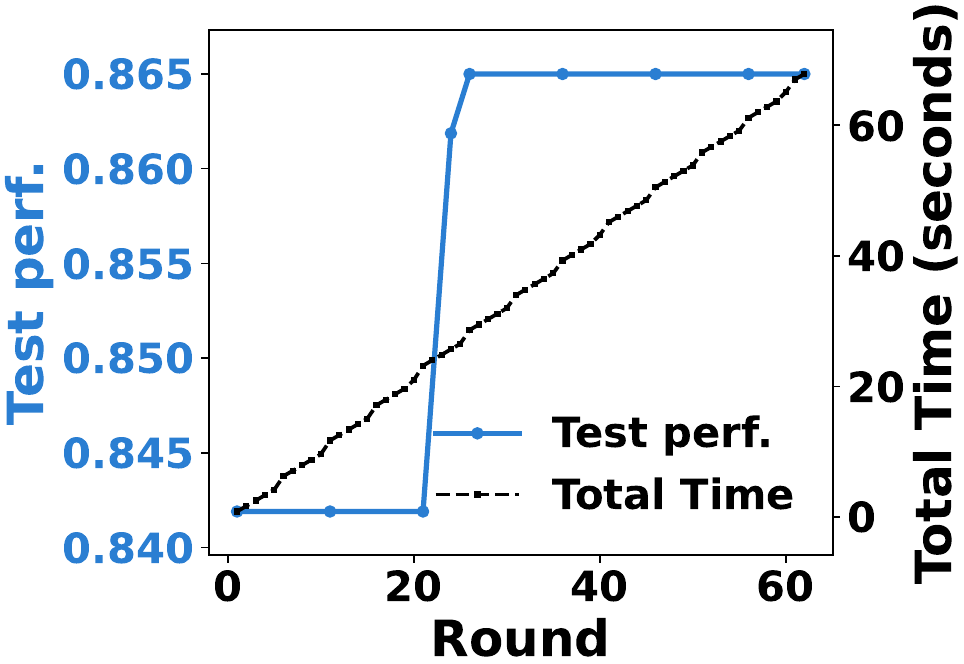}
\caption{PCMAC\_1}
\end{subfigure}
\hfill
\begin{subfigure}[t]{0.32\textwidth}
\centering
\includegraphics[width=\linewidth]{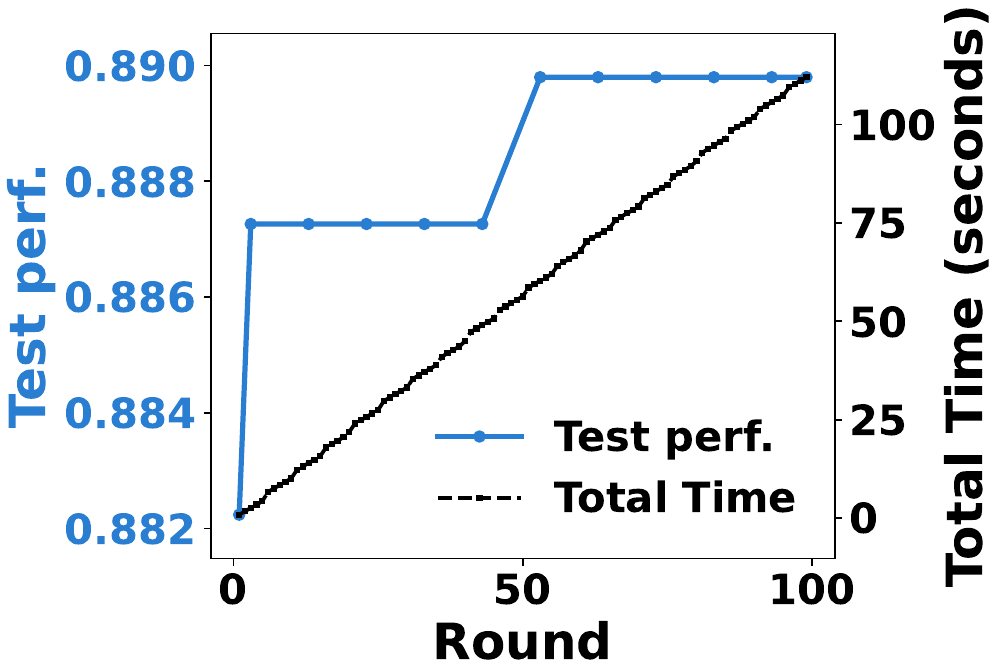}
\caption{BASEHOCK\_1}
\end{subfigure}

\vspace{0.1in}
\begin{subfigure}[t]{0.32\textwidth}
\centering
\includegraphics[width=\linewidth]{FIG/TIME_BNGcmc.pdf}
\caption{BNG(cmc)}
\end{subfigure}
\hfill
\begin{subfigure}[t]{0.32\textwidth}
\centering
\includegraphics[width=\linewidth]{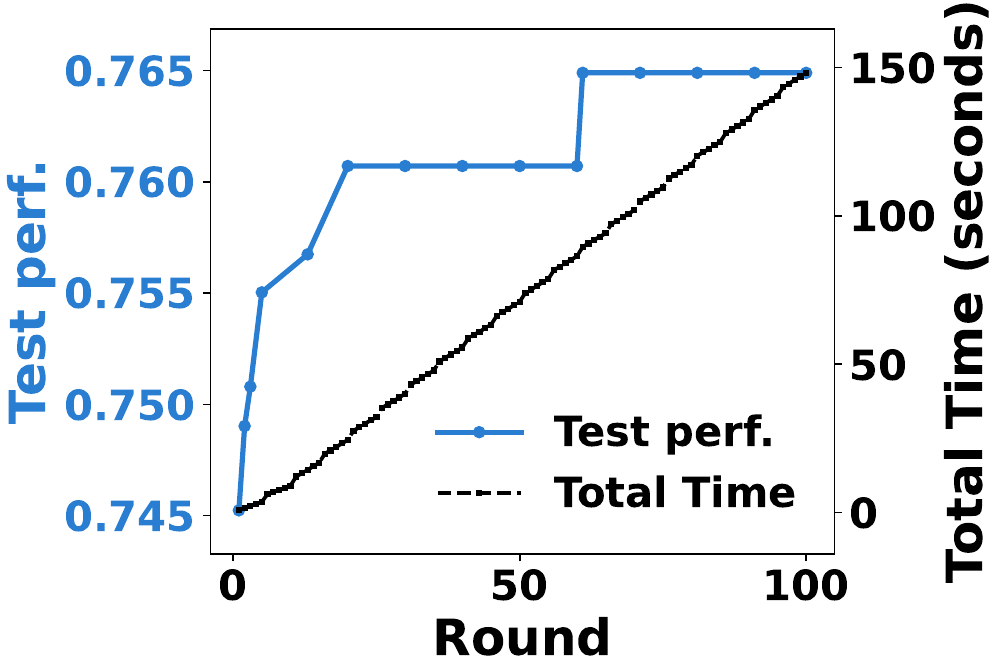}
\caption{electricity}
\end{subfigure}
\hfill
\begin{subfigure}[t]{0.32\textwidth}
\centering
\includegraphics[width=\linewidth]{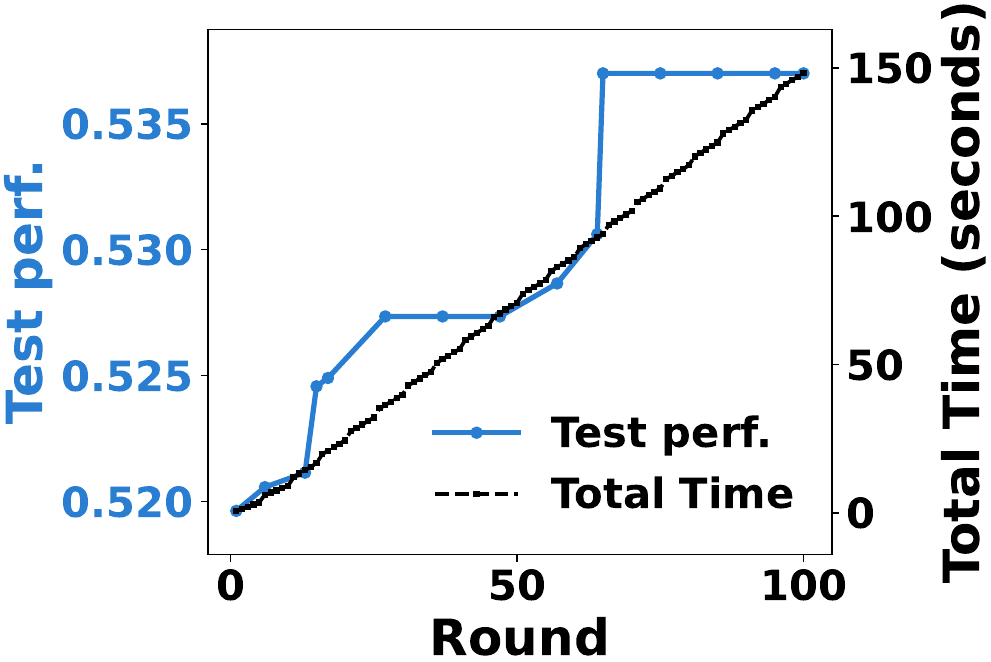}
\caption{FOREX\_audsgd-hour-High}
\end{subfigure}

\vspace{0.1in}

\begin{subfigure}[t]{0.32\textwidth}
\centering
\includegraphics[width=\linewidth]{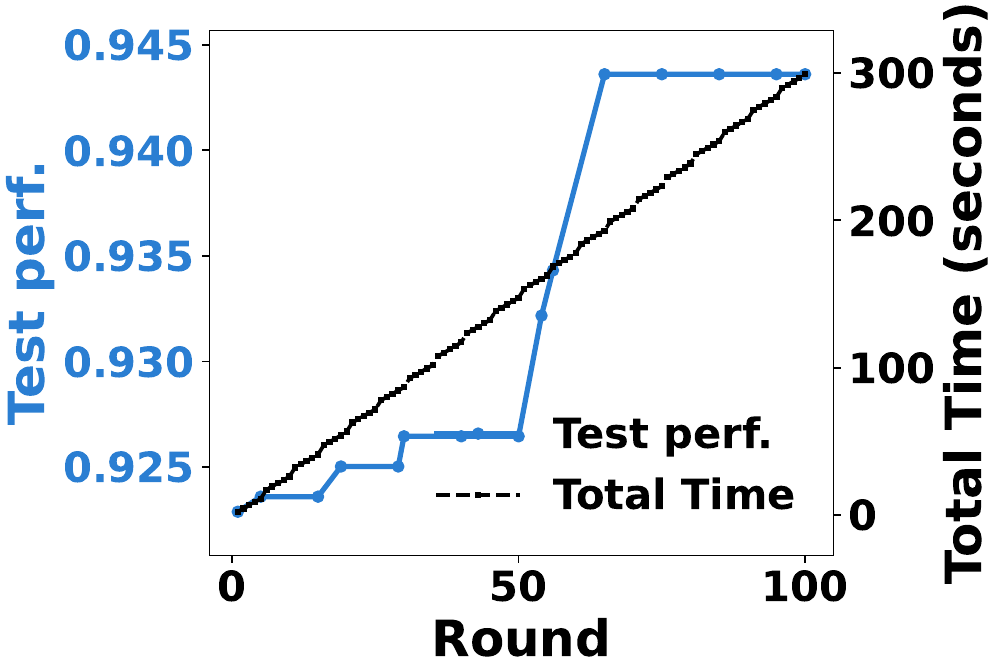}
\caption{gisette\_1}
\end{subfigure}
\hfill
\begin{subfigure}[t]{0.32\textwidth}
\centering
\includegraphics[width=\linewidth]{FIG/TIME_INNHotelsGroup.pdf}
\caption{INNHotelsGroup}
\end{subfigure}
\hfill
\begin{subfigure}[t]{0.32\textwidth}
\centering
\includegraphics[width=\linewidth]{FIG/TIME_jungle_chess_endgame.pdf}
\caption{jungle\_chess\_endgame}
\end{subfigure}

\vspace{0.1in}

\begin{subfigure}[t]{0.32\textwidth}
\centering
\includegraphics[width=\linewidth]{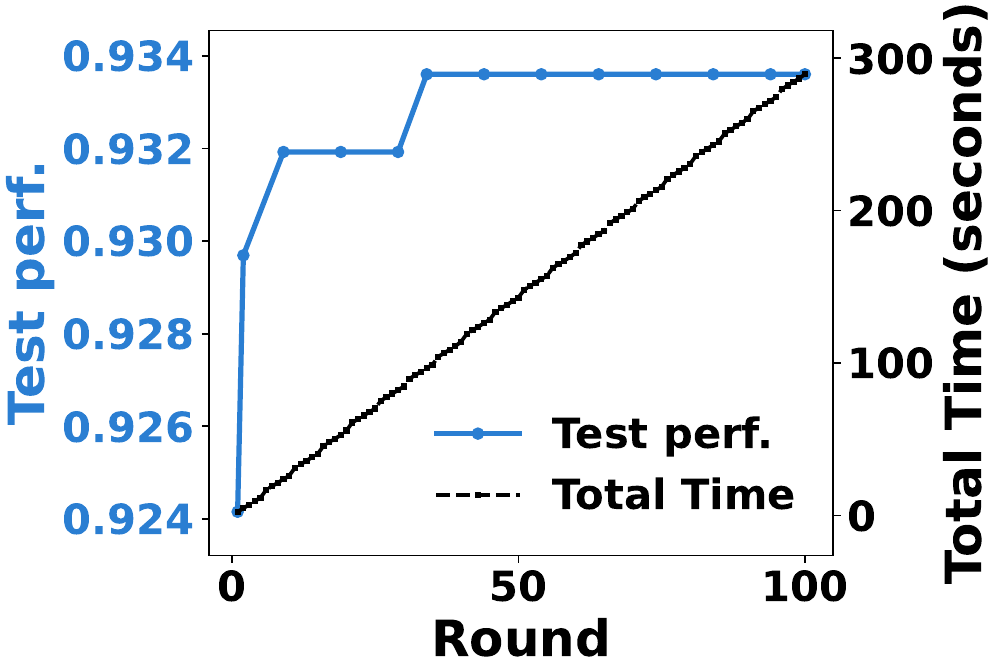}
\caption{nomao}
\end{subfigure}
\hfill
\begin{subfigure}[t]{0.32\textwidth}
\centering
\includegraphics[width=\linewidth]{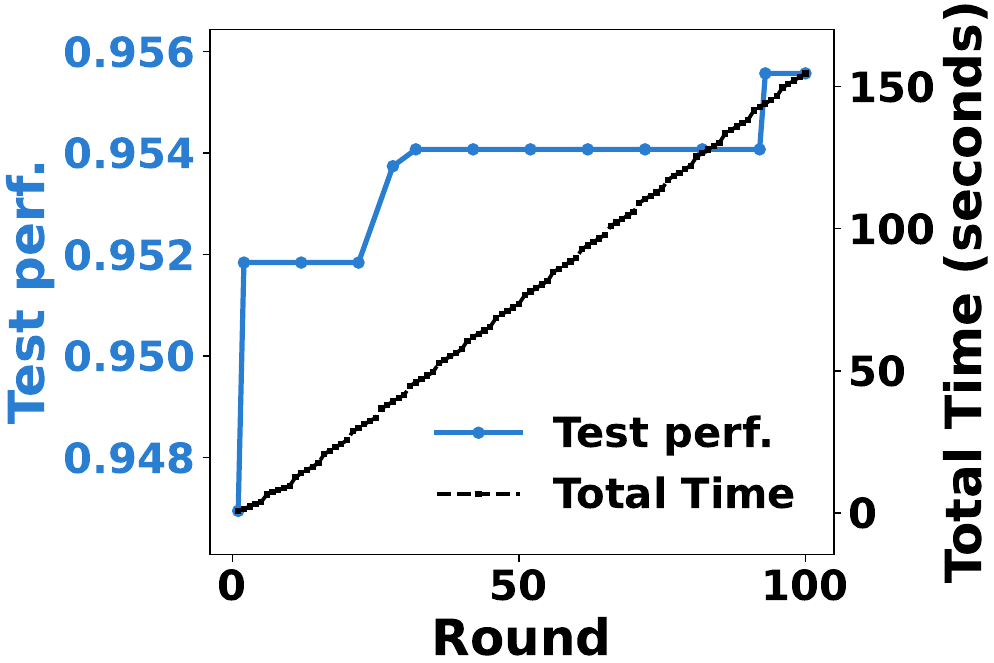}
\caption{SDSS17}
\end{subfigure}
\hfill
\begin{subfigure}[t]{0.32\textwidth}
\centering
\includegraphics[width=\linewidth]{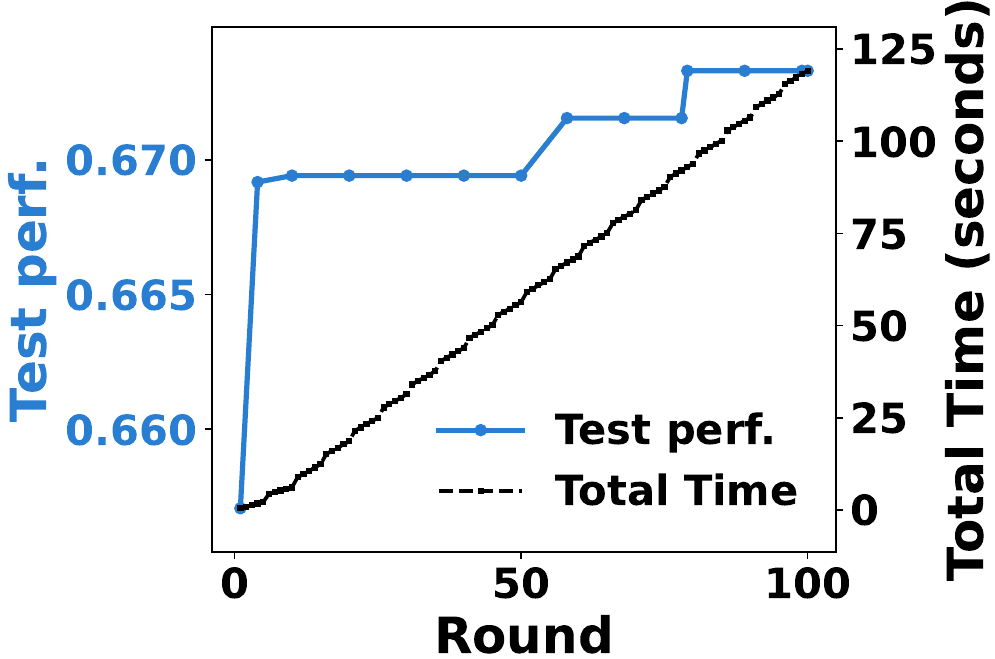}
\caption{shuttle}
\end{subfigure}

\vspace{0.1in}

\caption{\textbf{Running time} of \method (black) \textit{increases linearly} over optimization rounds, averaging approximately 1.4 seconds per round. Accordingly, \textbf{test performance} (blue) \textit{improves monotonically}, indicating that \method functions as an anytime method that can accommodate varying inference-time latency requirements.}
\label{fig:timegrid}

\end{figure*}

\begin{table}[t]
\centering
\caption{Dataset/setting combination count by \textbf{round of saturation} (TabICLv2 backbone)}
\label{tab:saturation}
\begin{tabular}{lccccc}
\toprule
\textbf{Dataset} & \textbf{0--25 rounds} & \textbf{25--50 rounds} & 
\textbf{50--75 rounds} & \textbf{75--100 rounds} & \textbf{Total Count} \\
\midrule
Original     & 6  & 9 & 7 & 2 & 24 \\
Augmentation & 3  & 6 & 5 & 2 & 16 \\
Noise        & 13 & 5 & 4 & 2 & 24 \\
\bottomrule
\end{tabular}
\end{table}
\clearpage

\section{Qualitative analysis of selected features: a Case Study}

\label{app:case_study}

\method supports interpretability. Unlike standard TFMs, which are largely black-box models, it provides feature-importance-based rankings of both features and samples, which can help explain decisions and guide the design of new features.

To demonstrate this capability, We evaluated \method's feature-selection mechanism on the UCI Communities and Crime Dataset \cite{communities_and_crime_183} to showcase \method’s interpretability in practical settings. The dataset contains 1,994 US communities (samples), 122 named attributes (features), and a continuous violent-crime-rate target. (\url{https://archive.ics.uci.edu/dataset/183/communities+and+crime})

During preprocessing, we removed 22 attributes with more than 50

The resulting \method ranking closely agreed with the creator-published univariate Pearson correlations despite being computed by entirely different mechanisms. \textbf{\method's top 4 features (\textit{PctKids2Par}, \textit{racepctblack}, \textit{racePctWhite}, \textit{PctIlleg}) were exactly the four attributes with the largest $|r|$ in the creator's analysis.} Of the Creator’s 10 strongest univariate predictors, 9 appeared in \method's top 20. We noticed that major criminological constructs like family-structure, divorce triplet, investment income, and racial-composition were represented in the top 20.

Beyond univariate correlation, we noticed \method also captures feature redundancy. Looking closely on where the two rankings diverge, the creator’s correlation table scored the 4 two-parent-household variants nearly identically ($|r| = 0.74/0.71/0.66/0.67$) because marginal correlation cannot see redundancy. In contrast, \method assigns sharply decreasing importance scores across this group (0.0089, 0.0039, 0.0027, and 0.0006), which reflects their diminishing marginal contributions. Similar behavior appeared in the poverty and economic block where \textit{pctWInvInc} was selected at rank 5 while its strongly correlated counterpart \textit{pctWPubAsst} was 84th. Conversely, several variables with relatively weak univariate correlations are promoted above stronger marginal predictors, suggesting that \method captures complementary information that is overlooked by univariate analyses.

From a practical perspective, the full analysis was a single unmodified run of the \method pipeline, which produces a complete ranked list with continuous scores. These scores naturally support feature-level interpretation while providing practitioners with a principled basis for understanding the model's feature preferences.

One limitation of this analysis is that it was conducted using a single random seed. A stability check across seeds could be done to add robustness to this analysis.

\end{document}